\documentclass[lettersize,journal]{IEEEtran}
\usepackage{amsmath,amsfonts}
\usepackage{algorithmic}
\usepackage{algorithm}
\usepackage{array}
\usepackage[caption=false,font=normalsize,labelfont=sf,textfont=sf]{subfig}
\usepackage{textcomp}
\usepackage{stfloats}
\usepackage{url}
\usepackage{verbatim}
\usepackage{graphicx}
\usepackage{cite}

\usepackage{times}
\usepackage{epsfig}
\usepackage{float}
\usepackage{amsmath}
\usepackage{amssymb}
\usepackage{bm}
\usepackage{tabu}
\usepackage{multirow}
\usepackage{makecell} 
\usepackage{CJKutf8}
\usepackage{xcolor}
\usepackage{colortbl}
\usepackage{multicol}

\usepackage{marvosym}  % letter logo

\usepackage[switch]{lineno}
\usepackage[pagebackref=true,breaklinks=true,colorlinks,bookmarks=false]{hyperref}
\usepackage{float}

\begin{document}

\title{A-SDM: Accelerating Stable Diffusion through Model Assembly  and Feature Inheritance Strategies}
\author{ Jinchao Zhu*\textsuperscript{\rm 1, 2, 3},Yuxuan Wang*\textsuperscript{\rm 3, 4},  Siyuan Pan\textsuperscript{\rm 3}, Pengfei Wan\textsuperscript{\rm 3}, Di Zhang\textsuperscript{\rm 3}, Gao Huang\Letter\textsuperscript{\rm 1}\\
\textsuperscript{\rm 1} Department of Automation, BNRist,  Tsinghua University,  Beijing, 100089, China\\
\textsuperscript{\rm 2} College of Software, Nankai University, Tianjin, 300000, China\\
\textsuperscript{\rm 3} Kuaishou Technology, Beijing, 100089, China\\
\textsuperscript{\rm 4} Department of Computing, Imperial College London, London, SW7 2AZ, the United Kingdom\\
\textsuperscript{\rm 3} School of Finance, Tianjin University of Finance and Economics, Tianjin, 300000, China\\
{\tt\small jczhu@mail.nankai.edu.cn, yuxuan.wang123@imperial.ac.uk, gaohuang@tsinghua.edu.cn, pansiyuan@kuaishou.com, wanpengfei@kuaishou.com, zhangdi08@kuaishou.com}

\thanks{* Joint first authors.} %*Equal contribution 
\thanks{This work is done when Jinchao Zhu and Yuxuan Wang are interns at Kuaishou Technology.}
}
% The paper headers
\markboth{Journal of \LaTeX\ Class Files,~Vol.~xx, No.~xx, June~2024}%
{Shell \MakeLowercase{\textit{et al.}}: A-SDM: Accelerating Stable Diffusion through Model Assembly and Feature Inheritance Strategies}

%\IEEEpubid{0000--0000/00\$00.00~\copyright~2021 IEEE}
% Remember, if you use this you must call \IEEEpubidadjcol in the second
% column for its text to clear the IEEEpubid mark.

\maketitle

\begin{abstract} 
The Stable Diffusion Model (SDM) is a prevalent and effective model for text-to-image (T2I) and image-to-image (I2I) generation. 
Despite various attempts at sampler optimization, model distillation, and network quantification, these approaches typically maintain the original network architecture. 
The extensive parameter scale and substantial computational demands have limited research into adjusting the model architecture. 
This study focuses on reducing redundant computation in SDM and optimizes the model through both tuning and tuning-free methods. 
1) For the tuning method, we design a model assembly strategy to reconstruct a lightweight model while preserving performance through distillation. 
Second, to mitigate performance loss due to pruning, we incorporate multi-expert conditional convolution (ME-CondConv) into compressed UNets to enhance network performance by increasing capacity without sacrificing speed.  
Third, we validate the effectiveness of the multi-UNet switching method for improving network speed. 
2) For the tuning-free method, we propose a feature inheritance strategy to accelerate inference by skipping local computations at the block, layer, or unit level within the network structure.
% Sampling mode
We also examine multiple sampling modes for feature inheritance at the time-step level.
% 6 Experiments  
Experiments demonstrate that both the proposed tuning and the tuning-free methods can improve the speed and performance of the SDM.
The lightweight model reconstructed by the model assembly strategy increases generation speed by $22.4\%$, while the feature inheritance strategy enhances the SDM generation speed by $40.0\%$.
% quantitative 
\end{abstract}


\section{Introduction}

% Diffusion
\IEEEPARstart{I}{n}  recent years, diffusion models have been rapidly developed and applied to various fields, such as text-to-image generation~\cite{2022-CVPR-LDM, 2022-arxiv-t2iclip, 2022-NIPS-t2iPhotorealistic,2022-ICML-GLIDE}, image-to-image translation~\cite{2021-ICCV-ILVR, 2022-SIGGRAPH-Palette}, image editing\cite{2022-ICLR-SDEdit, 2022-CVPR-Blended, 2023-CVPR-imagic}, image super-resolution~\cite{2023-TPAMI-SRRefine, 2022-NC-SRDiff}, data augmentation~\cite{2024-arxiv-LAKE-RED}, image segmentation\cite{2023-arxiv-camoDiffusion, 2023-arxiv-DMforCOD, 2024-arxiv-DiffSal}, reference-guided image generation~\cite{2023-CVPR-GLIGEN, 2023-ICCV-TF-ICON, 2024-AAAI-t2iadapter, 2023-ICCV-ConditionControl, 2023-arXiv-ProSpect}, personalized image generation~\cite{2022-arXiv-OneWord,2023-NIPS-photoswap,2023-ICCV-SVDiff,2023-CVPR-multiconcept, 2023-CVPR-DreamBooth}, text-to-video generation~\cite{2023-ICCV-t2v-zero,2023-CVPR-VideoFusion,2023-ICCV-TuneAVideo,2023-siggraph-RerenderAVideo}, and text-to-3D generation~\cite{2024-TNNLS-DMESH, 2023-CVPR-Magic3D, 2021-CVPR-Diffusion3DPoint, 2023-ICLR-DreamFusion}.
% Stable Diffusion
The Stable Diffusion Model (SDM) stands out as the foremost and widely recognized text-to-image (T2I) generation model. 
Its high-quality generation results have led to the adoption across various condition-guided visual tasks, including image-to-image (I2I) generation for style transformation, video generation, inpainting, etc.
SDM is a latent diffusion model~\cite{2022-CVPR-LDM} (LDM) for conditional image generation tasks, which improves computational efficiency by performing denoising processes in the latent space.

\begin{figure}[t]
\centering
\includegraphics[width=1\columnwidth]{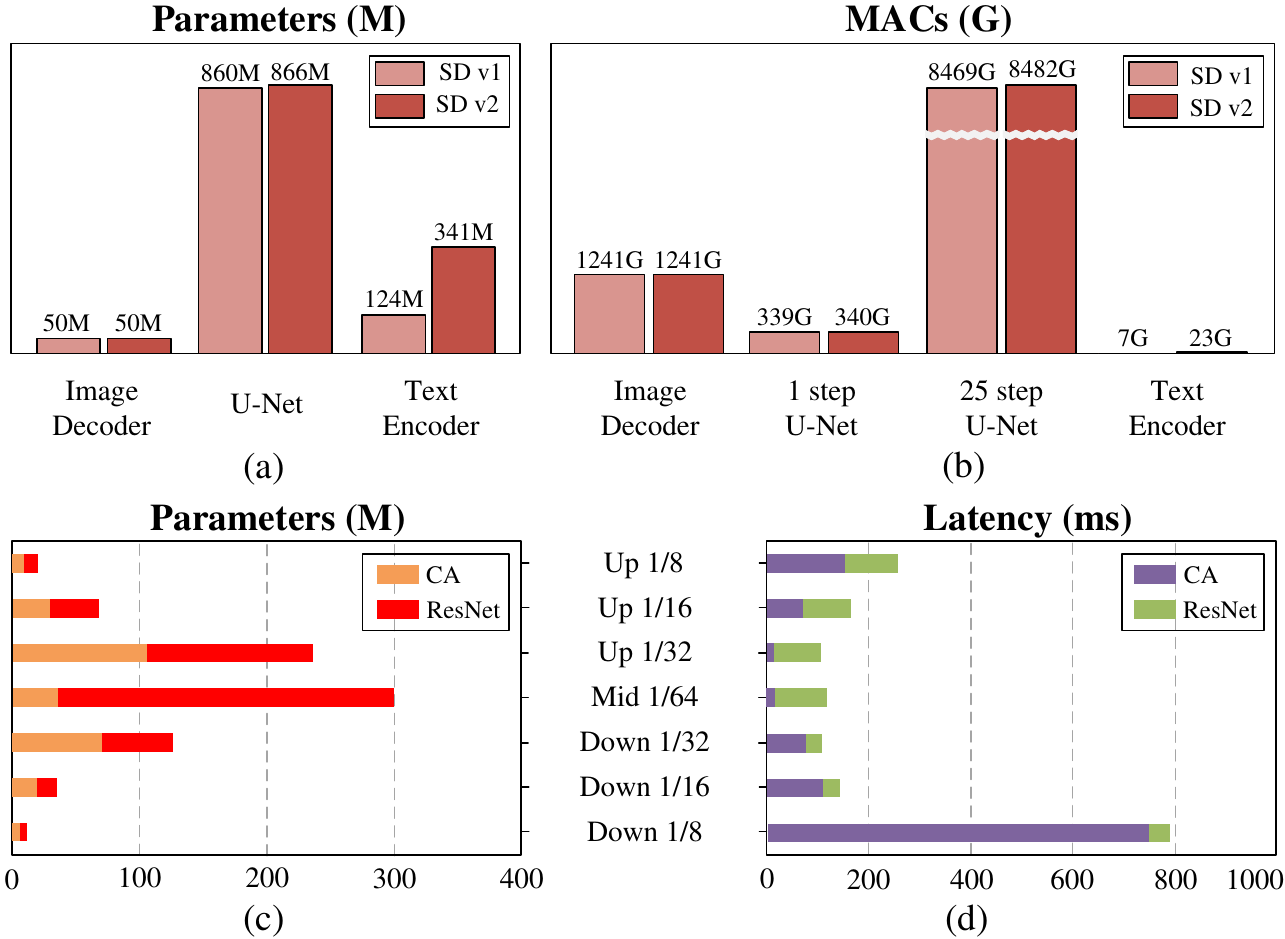}
\caption{(a) and (b) illustrate the parameters and computational requirements~\cite{2023-ICMLW-bksdm} of SDM v1 and v2. (c) and (d)  present analysis of the parameters and latency (iPhone 14 Pro, ms)~\cite{2023-nips-snapfusion} for cross-attention (CA) and ResNet blocks within the UNet of SDM.}
\label{01_SD_Analysis}
\end{figure}

% High computing cost for UNet
While latent space optimization is a significant improvement, the iterative denoising process of UNet within SDM still contributes to a considerable computational cost. 
It places a huge burden on computing resources, posing challenges for deploying SDM on mobile terminals. 
Fig.\ref{01_SD_Analysis} shows that the predominant consumption of computing resources occurs during the iteration of UNet. 

\IEEEpubidadjcol
% Previous methods
% train & non-train 
% structure & sampling optimization 
To mitigate this problem, a variety of optimization methods for SDM have been proposed, categorized into tuning and tuning-free methods.
The tuning methods encompass techniques such as pruning~\cite{2023-NIPS-StructurePruning, 2024-WACV-tokenfusion}, quantization~\cite{2023-NIPS-PTQD}, distillation~\cite{2023-ICMLW-bksdm, 2023-CVPR-ondistill, 2022-ICLR-progressivedistill}, and more.
Tuning-free methods involve optimizations like sampler optimization~\cite{2022-NIPS-dpmsolver}, token merging~\cite{2023-CVPRW-tome2, 2023-ICLR-tome1, 2024-WACV-tokenfusion}, etc.
These methods target network structure optimization and sampling optimization as the primary objectives.
Network structure optimization aims to reduce either the local~\cite{2023-ICMLW-bksdm, 2023-ICLR-tome1, 2023-CVPRW-tome2, 2024-WACV-tokenfusion, 2023-nips-snapfusion} or the overall~\cite{2023-CVPR-ondistill} computational consumption of the network. 
On the other hand, sampling optimization mainly focuses on reducing the number of network inference steps~\cite{2022-NIPS-dpmsolver,2022-ICLR-progressivedistill}.

% Lack of component function optimization
Although the above approaches accelerate the diffusion models from multiple perspectives, strategies for enhancing the UNet network component functions remain lacking. 
Therefore, this paper conducts experimental analysis on each UNet component and optimizes the network structure through both tuning and tuning-free approaches.

\begin{figure}[t]
\centering \includegraphics[width=1\columnwidth]{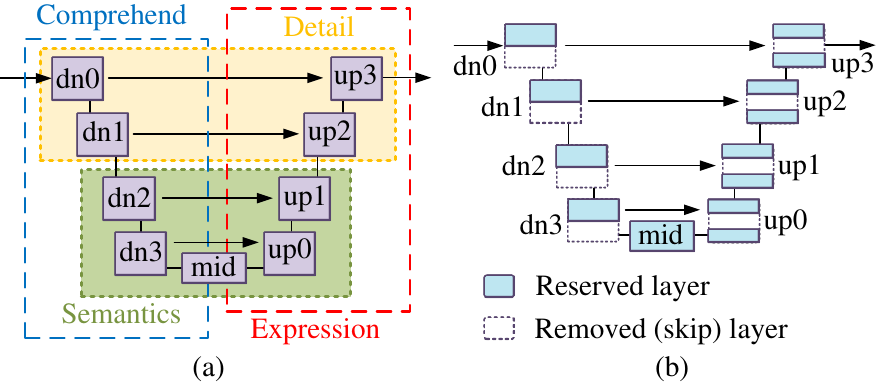}
\caption{(a) demonstrates the function of the different components within UNet. The encoder part (\textcolor[rgb]{0, 0.5, 1}{blue} box) is primarily responsible for understanding the input image while the decoder part (\textcolor{red}{red} box) handles the expressive reconstruction of the image. The shallow layers (\textcolor[rgb]{1, 0.8, 0}{yellow} block) focus on detail optimization while the deep layers (\textcolor[rgb]{0,0.5,0}{green} block) concentrate on semantic optimization. (b) presents a conceptual approach to network structure optimization. The blue squares indicate the layers within each block of the UNet that are retained. In the distillation method, the white dashed squares represent the blocks that have been removed. In the untrained feature inheritance strategy, these dashed squares denote the layers that skip internal calculations.}
\label{02_SD_Plaincut}
\end{figure}

% UNet
Originally, UNet was devised for medical image segmentation tasks \cite{2015-MICCAI-UNet}. It employs encoding and decoding structures to comprehend input images across multiple scales and derive segmentation outcomes. 
In the diffusion generation task, as illustrated in Fig.~\ref{02_SD_Plaincut} (a), the UNet retains the fundamental U-shape structure. 
Here, the encoder comprehends and interprets the input data, while the decoder primarily concentrates on generating and expressing image content. 
Shallow layers prioritize the embellishment of detail, while deeper layers focus on semantic modification and updating.

Based on the experience, this study delves into the optimization strategy of UNet components on two core issues: \textit{1) how to accurately remove the redundant parts of the standard architecture and improve performance through efficient distillation tuning} and \textit{2) how to achieve tuning-free acceleration by skipping negligible computing units, layers, or blocks.} 

To address the first issue, we adopt a model assembly strategy. 
First, we employ a straightforward distillation approach, evenly pruning each block in the network by one layer to obtain the compressed model, as depicted in Fig.~\ref{02_SD_Plaincut} (b). 
According to Fig.~\ref{01_SD_Analysis} (c) and (d), we observe that the shallow blocks of dn0, dn1, up2, and up3 exhibit fewer parameters and higher latency, making them suitable candidates for distillation.
Second, we merge the shallow layer blocks of the compressed model with the deep layer blocks of the original SDM to create the reconstructed model which undergoes a second round of distillation. 
Additionally, for small model optimization, we integrate conditional convolution (CondConv)~\cite{2019-NIPS-CondConv} into the pruned blocks to enhance the network's capacity. 
Furthermore, in terms of model assembly optimization, we explore a multi-UNet switching approach, using compressed SDM and the original SDM during the early and late sampling period, respectively, to achieve acceleration.
The method yields promising experimental results, demonstrating its effectiveness in optimizing the UNet components.

For the second issue, our goal is to develop a tuning-free method to substitute the distillation method, enabling the skipping of trivial layer calculations, as shown in Fig~\ref{02_SD_Plaincut} (b).
Given that the UNet performs iterative denoising processes, the features in adjacent UNet inference processes exhibit similarity.
Leveraging the observation, we propose a feature inheritance strategy to skip insignificant calculations in the current step by inheriting features from the previous step.
Furthermore, considering the characteristics of UNet components, we explore local skip designs for shallow layers, deep layers, encoders, and decoders, respectively.
Besides, we investigate multiple sampling modes for feature inheritance at the time-step level, to determine the optimal approach for enhancing inference efficiency while preserving model performance.
% various feature inheritance schemes, including early, middle, late, and global strategies, to determine the optimal approach for enhancing inference efficiency while preserving model performance.
% We also examine multiple sampling modes for feature inheritance at the time-step level.

\begin{itemize}     
\item We present a model assembly strategy to reconstruct an efficient and semantically stable lightweight model by fine-tuning.
% We propose a general tuning-free feature inheritance strategy to achieve computation skipping in non-critical parts.
% We identify and prune the computational redundancy parts within the network to enhance efficiency. We further improve the network performance by employing a model assembly strategy.
To preserve the performance of the pruned model, we introduce multi-expert conditional convolution (ME-CondConv) to increase model capacity.
Besides, we devise a multi-UNet switching approach to collaborate models of varying capacities for faster generation.
%, enabling faster generation without sacrificing performance.

\item We propose a general tuning-free feature inheritance strategy to achieve computation skipping in non-critical parts. 
The feature inheritance strategies on multiple structures are explored, including block-level, layer-level, unit-level, and concurrent feature inheritance.
%Various feature inheritance strategies are explored, 

\item  Qualitative and quantitative experiments demonstrate that both the proposed tuning and the tuning-free methods are effective and efficient.
% The model assembly method accelerates the UNet component by $22\%$  
%compared to the original SDM. 
% The feature inheritance strategy further boosts the inference speed by $40\% $without additional training.
The lightweight model reconstructed by the model assembly strategy increases generation speed by $22.4\%$ compared to the original SDM.
The feature inheritance strategy enhances the SDM generation speed by $40.0\%$.
\end{itemize}

\section{Related Work}
In recent years, high-quality image generation has developed rapidly.
GANs~\cite{GAN, cGAN, cycleGAN, styleGAN, Pix2Pix} and VAEs~\cite{VAE, cVAE, AVB, NVAE, VQVAE} are early mainstream generation methods, but these models are unstable and difficult to control semantics information.
With the emergence of the diffusion model, a new round of research on image generation models has been initiated. Models like~\cite{2021-NIPS-DMBeatGAN, 2020-NIPS-DPM, 2021-ICLR-DDIM} have attracted significant attention due to their stable high-fidelity image generation capability.

\subsection{T2I Diffusion Models} 

The diffusion model can achieve accurate and high-quality T2I image generation by leveraging the pre-trained language model~\cite{2021-ICML-CLIP}. 
Typical T2I models include GLIDE~\cite{2022-ICML-GLIDE}, Imagen~\cite{2022-NIPS-ImageN}, DALL·E-2~\cite{2022-arXiv-DallE2}, DiT~\cite{2023-ICCV-DiT} and SDM~\cite{2022-CVPR-LDM}.
Among these, SDM stands out as an effective and open-source method, garnering widespread adoption and research interest.
Our research will build upon the SDM due to its stable and realistic image generation potential. 
To improve computing efficiency, SDM achieves high-speed diffusion generation in a low-dimensional latent space by employing a pixel-space autoencoder.
Within the latent space, a UNet~\cite{2015-MICCAI-UNet} architecture is adopted for efficient generation. Originally designed for medical image segmentation, UNet offers multi-scale feature processing capabilities owing to its U-shaped structure. This enables the generated images to exhibit high-quality details and semantics.
The UNet architecture is augmented with Transformer units at each layer to enhance global awareness and facilitate cross-modal interaction.
Although the SDM achieves latent space generation, the cyclic denoising mechanism of diffusion leads to slow inference. 
In this paper, we refer to common network design techniques used in segmentation tasks to optimize UNet within SDM. 
Segmentation models like TransUnet~\cite{2021-arXiv-TransUNet}, Deeplab~\cite{2018-TPAMI-DeepLab}, and CPD~\cite{2019-CVPR-CPD} simplify shallow computations and concentrate computing resources at deeper layers. 
As such, we believe that the breakthrough point of acceleration lies in compressing the shallow layers of UNet with fewer parameters.

\subsection{Sampling Acceleration}  % Diffusion Solver Optimization
% The goal of these methods is to reduce sampling steps and improve the quality of generation by improving the diffusion solver.
DDIM~\cite{2021-ICLR-DDIM} is proposed to generalize DDPM~\cite{2020-NIPS-DPM} via a class of non-Markovian diffusion processes, which effectively reduces sampling steps.
Subsequently, more fast high-order solvers~\cite{2022-NIPS-dpmsolver, 2022-arXiv-dpm++, 2022-ICLR-Pseudo, 2023-ICLR-fast, 2022-ICLR-analitic-dpm} have been designed for sampling acceleration.
DPM-Solver~\cite{2022-NIPS-dpmsolver}, a dedicated higher-order solver for diffusion ODEs, achieves convergence order guarantee and fast sampling.
Additionally, consistency models~\cite{2023-arXiv-consistency} can generate high-quality samples by directly mapping noise to data and enable the generation of one- and few-step sampling.

\subsection{Distillation Acceleration} 
% progressive
The progressive distillation strategy~\cite{2022-ICLR-progressivedistill} is proposed to distill the behavior of an N-step DDIM sampler into a new model with N/2 steps, minimizing degradation in sample quality. This iterative method allows for further reduction of sampling steps while maintaining quality.
% on distillation
Building upon the principles of progressive distillation, Meng \textit{et al.}~\cite{2023-CVPR-ondistill} further devised a two-stage distillation technique. This approach refines the evaluation process for diffusion models, consolidating the traditional model evaluation of two into one. 
BOOT~\cite{2023-arxiv-BOOT} introduced an efficient data-free distillation algorithm, avoiding expensive training procedures.
The methods described above maintain the UNet structure. In the subsequent discussion, we will introduce distillation approaches that involve modifying the architecture.
% early stop

\subsection{Architecture Compression}% Architecture Efficiency 

Small SDM~\cite{2023-http-SmallSD} first explores the UNet compression strategy of uniform layer pruning by distillation, illustrated in Fig.~\ref{02_SD_Plaincut} (b). 
%Subsequently, a distillation method is employed to preserve the performance of the pruned model.
Building upon this, BK-SDM~\cite{2023-ICMLW-bksdm} further investigates the compression strategy introduced by small SDM. Specifically, it extends the pruning to deeper blocks of SDM, resulting in even smaller models. 
Additionally, BK-SDM introduces a high-quality small-scale training dataset to significantly reduce distillation time.
% to accelerate the distillation training process. 
Moreover, SnapFusion~\cite{2023-nips-snapfusion} proposes an evaluation mechanism leveraging CLIP~\cite{2021-ICML-CLIP} and latency metrics to evolve the UNet architecture during training.
% Specifically, we propose an efficient UNet by identifying the redundancy of the original model and reducing the computation of the image decoder via data distillation.
% deepcache: efficient structure evolving in SnapFusion
% bksdm: SnapFusion [31] achieves an efficient U-Net for SDMs through architecture evolution and step distillation 
Chen \textit{et al.}~\cite{2023-CVPRW-SpeedIsAllYouNeed} employs a range of implementation optimizations, including flash attention and Winograd convolution to expedite the diffusion model.
SimpleDiffusion~\cite{2023-ICML-SimpleDiffusion} proposes a U-ViT architecture, combining a U-Net with a Transformer backbone. The method demonstrates that high-resolution image generation can be achieved while maintaining simplicity in model structure.
Building on the concept of U-ViT, MobileDiffusion~\cite{2023-arXiv-MobileDiffusion} reconfigures the distribution of ResNet blocks and Transformer blocks, relocating the computationally intensive Transformer block to the low-pixel parts.
Specifically, Transformer blocks in the shallow layer ($64\times64$) and self-attention in the middle layer ($32\times32$) are removed.

% mobile sd
% [15] simple diffusion: End-to-end diffusion for high resolution images.
% [15] proposed the UViT architecture for training diffusion models on  high-resolution images.
% They demonstrated that superior quality and efficiency could be achieved by scaling transformer blocks at the low-resolution segment of the UNet. 

% [20] bksdm 
% [23] snapfusion Text-to-image diffusion model on mobile devices within two seconds 
% Meanwhile, [23] presented an efficient architecture search method.    They trained a UNet with redundant blocks using robust training techniques [17, 58] and then pruned certain blocks based on  metrics, resulting in an architecture suitable for distillation. 

% [17] Deep networks with stochastic depth 
% [58] Slimmable neural networks 
% 16 26 49 
% early stop

% -> faster
% snapfusion
% -> deepcache
% early stop
% -> bksdm
% diff-pruning

\subsection{Quantization Acceleration} 

To address the issue of slow iterative inference of diffusion models, several quantization methods~\cite{2023-ICCV-Q-Diffusion, 2023-NIPS-PTQD, 2022-http-A-SDM, 2023-http-android} have been proposed.
Q-Diffusion~\cite{2023-ICCV-Q-Diffusion} utilizes time step-aware calibration and split shortcut quantization to accelerate the generation process.
PTQD~\cite{2023-NIPS-PTQD} offers a unified formulation for quantization noise and diffusion perturbed noise. The method disentangles the quantization noise into correlated and uncorrelated components relative to its full-precision counterpart. Subsequently, it corrects the correlated component by estimating the correlation coefficient.
% bk-sdm
% [22] World’s first on-device demonstration of stable diffusion on an android phone
% [30]Q-diffusion: Quantizing diffusion models
% [68]  Accelerate stable diffusion with intel  neural compressor.
% []PTQD: Accurate Post-Training Quantization for  Diffusion Models

\subsection{Untrained Acceleration} 

Training-free acceleration approaches usually employ token merging~\cite{2023-CVPRW-tome2, 2023-ICLR-tome1, 2024-WACV-tokenfusion} and early stop sampling methods~\cite{2023-ICCV-AutoDiffusion, 2023-arXiv-DeeDiff} to reduce generation time.
ToMe~\cite{2023-CVPRW-tome2, 2023-ICLR-tome1} merges redundant tokens in the shallow layers for efficient computation, ensuring high-quality image generation without additional training.
ToFu~\cite{2024-WACV-tokenfusion} integrates the advantages of token pruning and merging, dynamically adjusting to each layer's properties for optimal performance.
AutoDiffusion~\cite{2023-ICCV-AutoDiffusion} presents a unified, training-free framework that explores optimal time steps and architectures within the diffusion model's search space, effectively enhancing sampling speed.
Similarly, DeeDiff~\cite{2023-arXiv-DeeDiff} introduces a timestep-aware uncertainty estimation module to estimate the prediction uncertainty of each layer. This uncertainty serves as a signal for determining when to terminate the inference process.

\section{Preliminary}

\subsection{Diffusion Principle}

Diffusion model~\cite{2020-NIPS-DPM} comprises forward  and reverse processes.
The forward process progressively perturbs the original image $x_0 \sim q(x_0)$ at time $t$ by injecting Gaussian noise with variance $\beta_t\in(0, 1)$ until it converges to isotropic Gaussian distribution: 
\begin{equation}
\begin{split}
q(x_t|x_{t-1}) := \mathcal{N}(x_t; \sqrt{1-\beta_t}x_{t-1}, \beta_t\textbf{I}).
\end{split}
\end{equation}

In the reverse process, the objective is to gradually revert random noise $x_T\sim \mathcal{N}(0,\textbf{I})$ to the expected distribution $x_0$ by removing noise. 
Training the reverse process ideally involves learning the exact inversion of the forward pass. However, directly estimating $q(x_{t-1}|x_t)$ from the dataset is challenging due to its dependence on the entire dataset. Instead, diffusion models utilize a deep neural network model $p_\theta$ to approximate this conditional distribution.
For the $t$ th reverse step, the sampling process involves calculating:
\begin{equation}
\begin{aligned}
&x_{t-1} \sim p_\theta(x_{t-1}|x_t) = \\
&\mathcal{N}\left(x_{t-1};\frac{1}{\sqrt{\alpha_t}}\left(x_t-\frac{\beta_t}{\sqrt{1-\bar{\alpha}_t}}\epsilon_{\theta}(x_t, t)\right), \beta_t\textbf{I}\right),
\end{aligned}
\end{equation}
where $\alpha_t = 1-\beta_t$ and $\bar{\alpha}_t=\prod_{s=1}^T{\alpha_s}$. 

\subsection{UNet Architecture}
To facilitate the expression in the following text, we categorize the entire UNet architecture into three levels: block, layer, and unit.
% As shown in Fig.3, 
In the SDM framework, the standard UNet comprises 4 down blocks, 4 up blocks, and 1 middle block.
Each down block consists of 2 layers, while each up block comprises 3 layers. Every layer includes a ResNet unit and a Transformer unit.
Notably, the deepest down block and up block do not contain Transformer units.
Additionally, the middle block consists of 2 ResNet units and 1 Transformer unit.

\begin{figure}[ht]
\centering
\includegraphics[width=1\columnwidth]{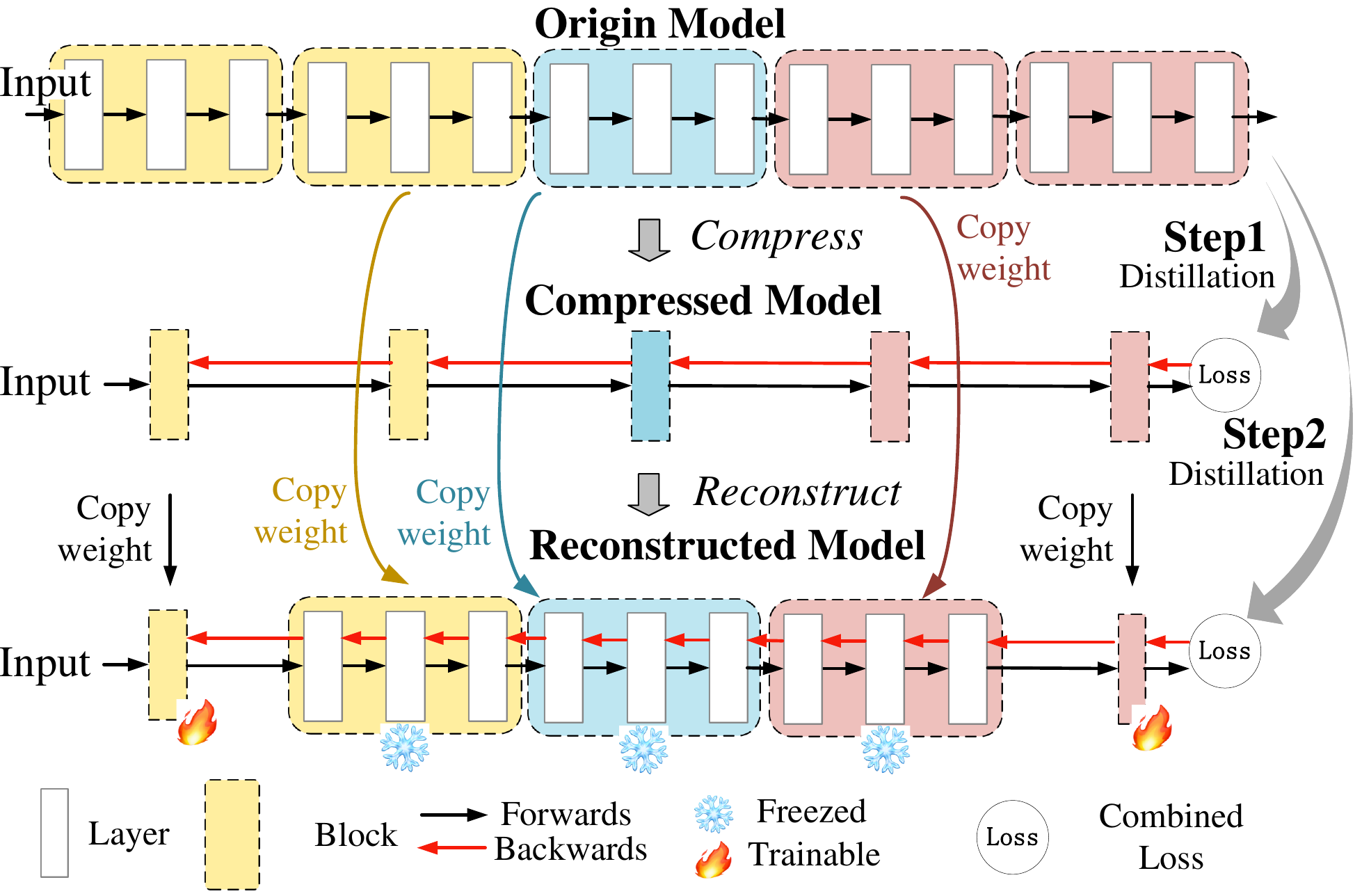}
\caption{Macro model assembly process. The first step is to compress the original model into a compressed model through distillation. The second step combines the original model's middle part with the compressed model's two sides to form a reconstructed model. The original model is then used as a teacher to distill the reconstructed model further.
}
\label{03_Incubation_Macro}
\end{figure}

\begin{figure}[t]
\centering
\includegraphics[width=1\columnwidth]{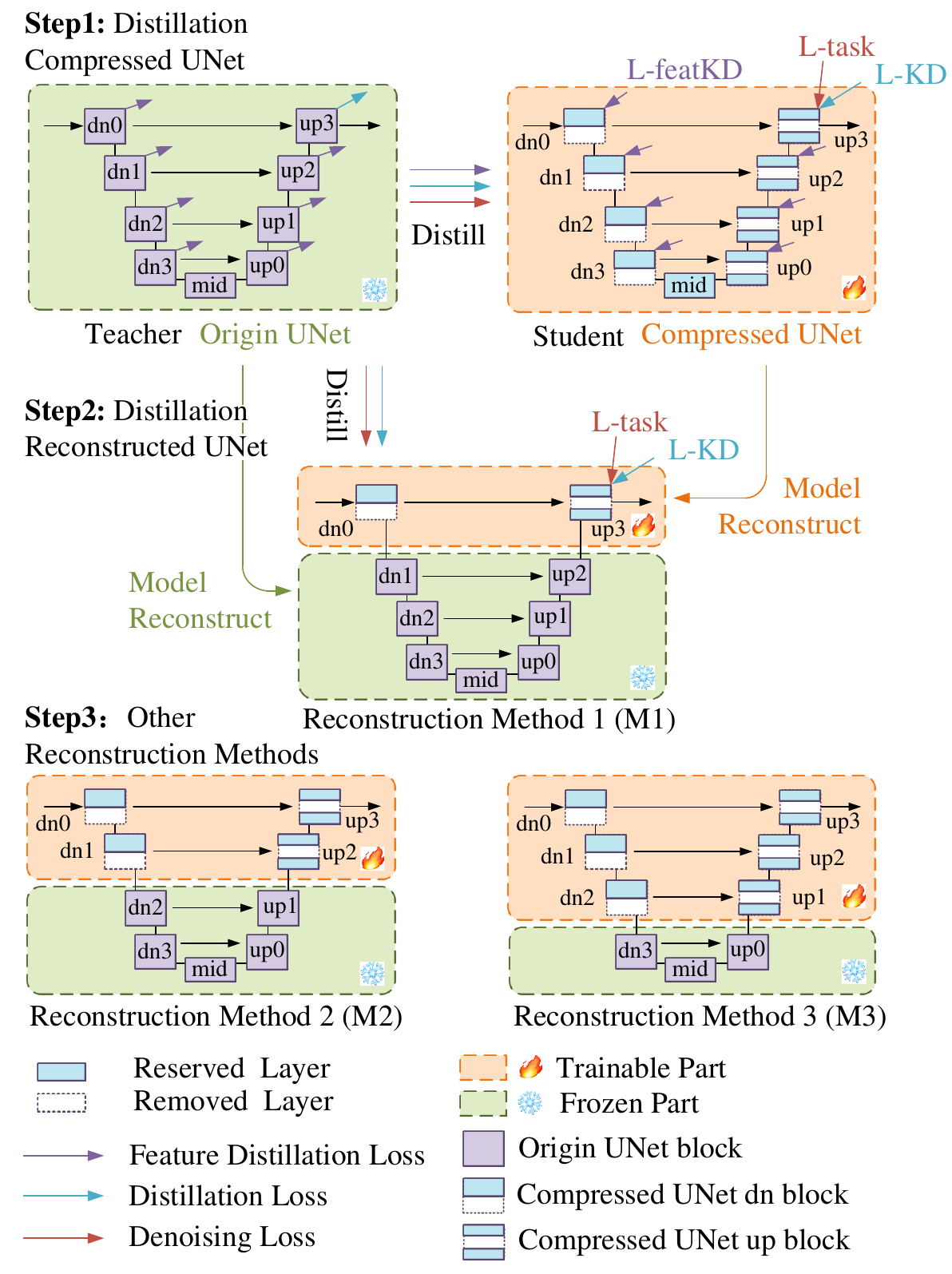}
\caption{Specific SDM model assembly process. \textbf{Step 1}: Distillation of the student compressed model. \textbf{Step 2}: The compressed UNet is merged with the original UNet to obtain the reconstructed UNet, which is then distilled again. Deep parts of the reconstructed UNet are frozen during training to ensure stable semantic generation. \textbf{Step 3}: Explore different combinations and distill the reconstructed UNet again. \textit{L-task} indicates the supervision of the denoising task. \textit{L-KD} represents distillation supervision at the UNet output position. \textit{L-featKD} denotes the distillation supervision of each block output. Loss settings refer to~\cite{2023-ICMLW-bksdm}.
}
\label{04_Incubation_Micro}
\end{figure}

\section{Model Assembly Strategy}
\label{section_training}

\subsection{Model Compression and Reconstruction}
\label{model compression}

Tuning Stable Diffusion Models typically requires substantial computational resources, deterring researchers from exploring model architecture reconstruction.
To tackle this challenge, we introduce a model assembly strategy aimed at enabling training with relatively fewer computational demands. 
Fig.\ref{03_Incubation_Macro} showcases our training approach from a simplified macro perspective.
Initially, we distill a compressed model, which is a smaller model that removes certain layers in each block. 
Subsequently, we integrate the components of the compressed model with those of the original model to create a new student model.
Notably, the original model components of the combined model remain frozen, effectively alleviating the computational burden during subsequent knowledge distillation tuning.

Fig.~\ref{04_Incubation_Micro} depicts the model assembly process of the UNet model within the specific SDM framework. 
In \textbf{step 1}, the original UNet serves as the teacher to distill the compressed student UNet, referred to as the Base model~\cite{2023-ICMLW-bksdm}.
This compression involves pruning one layer from each block of the original UNet, except for the middle block. 
In \textbf{step 2}, the deep layer part (green) of the original UNet is combined with the shallow part (orange) of the compressed UNet to obtain the reconstructed UNet.
The reconstructed UNet serves as the student model for a new round of distillation, where the deep layer part from the original UNet is frozen.
In \textbf{step 3}, two alternative combinations are explored for the second distillation step, resulting in three reconstructed models (M1, M2, and M3).

Experimental results demonstrate that reconstruction method 2 (M2 Model) is the most effective.
Furthermore, freezing the deep layer parameters of the reconstructed model leads to superior results compared to not freezing them.

\textbf{Discussion}: The shallow blocks of UNet have fewer parameters and higher latency (Fig.~\ref{01_SD_Analysis} (c), (d)), meaning that these locations are more conducive to model acceleration and fine-tuning learning.
Experiments show that they are more suitable for model reconstruction.
Since the deep blocks contains numerous parameters (Fig.~\ref{01_SD_Analysis} (c)) and the processing feature scale of the corresponding position is small, fine-tuning with a small-scale training dataset is challenging and the acceleration benefits obtained by model reconstruction are limited.
Therefore, freezing operation is beneficial for semantic stability and more cost effective.

% The shallow part consumes more time (Fig. \ref{01_SD_Analysis} (d)), while maintaining the deep part of the reconstructed model unchanged does not result in slower inference. 
%Since the deep part contains numerous parameters (Figure \ref{01_SD_Analysis} (c)), freezing it preserves the accuracy of semantic generation.
%Moreover, assembling the shallow layer with fewer parameters proves more effective. 

\subsection{Multi-Expert Conditional Convolution}

\begin{figure}[t]  % htb
\centering
\includegraphics[width=0.9\columnwidth]{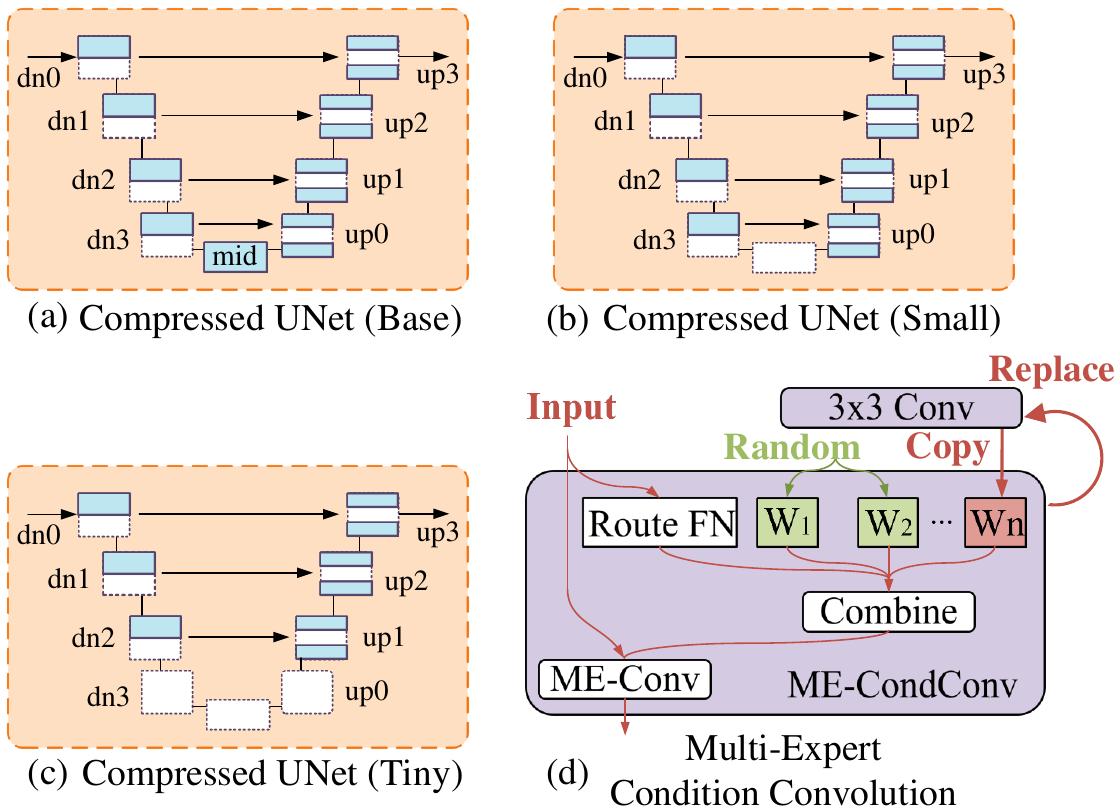}
\caption{Base UNet (a), Small UNet (b), and Tiny UNet (c) are the three compressed UNet structures proposed in~\cite{2023-ICMLW-bksdm}.  The white squares represent the layers have been pruned.  (d) illustrates ME-CondConv, which is adopted to expand the capacity of compressed UNets.
}
\label{05_ME-CondConv}
\end{figure}

In addition to the uniform layer removal scheme (Fig.~\ref{05_ME-CondConv} (a) Base model) and assembled model (Fig.~\ref{04_Incubation_Micro}), we further devise smaller models inspired by \cite{2023-ICMLW-bksdm}. 
Following the methodology, we first remove modules to construct the compressed models, namely the Base, Small, and Tiny.
The Small model removes the middle block (Fig.~\ref{05_ME-CondConv} (b)) from the Base, while the Tiny model (Fig.~\ref{05_ME-CondConv} (c)) additionally eliminates the deepest down and up blocks.
The distillation method employed to obtain the Base, Small, and Tiny models mirrors that of~\cite{2023-ICMLW-bksdm}.

However, the substantially reduced network capacity leads to a significant deterioration in model performance.
To address this challenge, we leverage multi-expert conditional convolution (ME-CondConv)~\cite{2019-NIPS-CondConv} to enhance network capacity without increasing inference time.  
As depicted in Fig.~\ref{05_ME-CondConv} (d), we extract the weights of $3\times3$ convolution in the compressed model as expert $W_{n}$, while other experts $W_{1}$,... , $W_{n-1}$ are initialized randomly. 
Dynamic routing is then employed to dynamically allocate weights to these experts based on input features, integrating the kernels of multiple experts for feature processing.
This process results in a new ME-CondConv from the original $3\times3$ convolution. Subsequently, we replace each $3\times3$ convolution in the compressed model with ME-CondConv accordingly.
Finally, the new Base, Small, and Tiny models with ME-CondConv are distilled following the approach outlined in~\cite{2023-ICMLW-bksdm}.

ME-CondConv effectively enhances the performance of the Small and Tiny models.
However, its optimization effect is less pronounced for the Base model, which possesses relatively large parameters. 
It is also unsuitable for the reconstructed models (M1, M2, M3) with more parameters depicted in Fig.~\ref{04_Incubation_Micro}.

\subsection{Multi-UNet Switching Approach} 
\label{multi-unet}

T2I task aims to generate target images from noise, whereas early-stage images of the denoising process tend to be relatively rough and lack detailed information. 
Conversely, the later stage requires polishing to enhance image quality.
This insight motivates us to design multiple UNets for a 25-step T2I process.
Initially, a compressed UNet (Base) is employed to rapidly generate the prototype in the initial stage (first 10 steps). 
Subsequently, the original UNet is utilized in the later stage (last 15 steps) for image optimization. 
The experimental results in Tab.~\ref{Tab-strategy} demonstrate that this strategy yields the best performance. 
It enhances speed and carries implications for our subsequent feature inheritance strategy of Sec. \ref{SamplingMode}.

\section{Feature Inheritance Strategy} 

Fig.~\ref{02_SD_Plaincut} (b) depicts a straightforward network acceleration scheme where the computation of one layer per block in the UNet is uniformly reduced.
In Sec.~\ref{section_training},  we proposed a \textit{tuning} strategy to distill the layer-pruned compressed model.
In this section, we introduce an \textit{tuning-free} feature inheritance strategy designed to skip certain layer calculations based on the circular inference mechanism of diffusion models.

\begin{figure}[t]
\centering
\includegraphics[width=1\columnwidth]{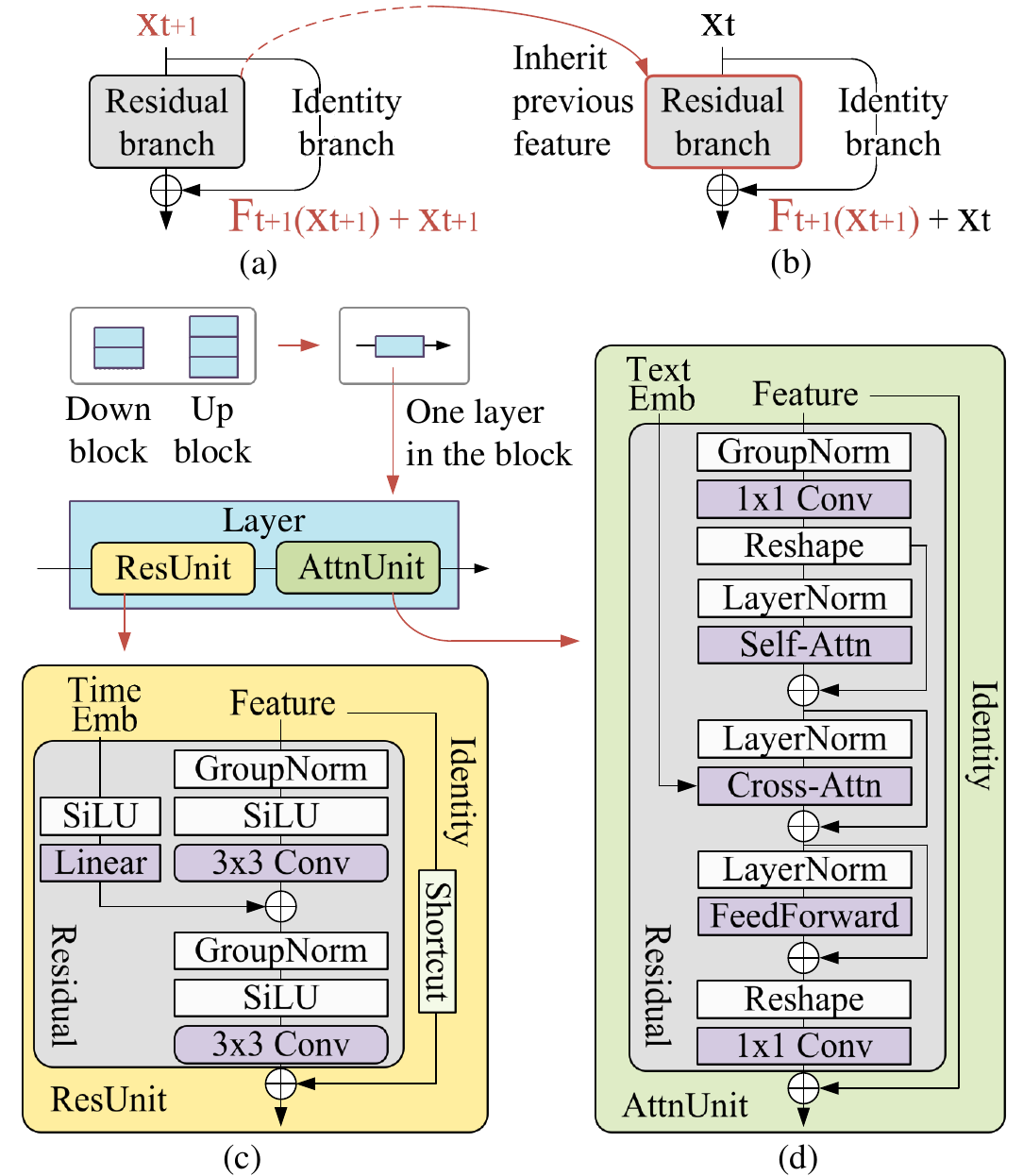}
\caption{Feature inheritance principle. (a) Traditional residual structure. (b) Residual structure of feature inheritance. (c) ResNet unit. (d) Attention unit.
}
\label{07_FeatureInheritance}
\end{figure}

The diffusion model shares the same UNet to denoise the input noise in multiple rounds, typically involving 50 sampling steps in SDM.
However, some calculations in this process are redundant and inefficient. To address this, we intend to leverage the features from the corresponding positions in the previous step to replace the current step's calculations. It enables us to omit the redundant calculation process, thereby accelerating the overall procedure.
Fig.~\ref{07_FeatureInheritance} (a) illustrates the residual structure~\cite{2016-CVPR-ResNet} commonly used in UNet.
For the input $x_{t+1}$ at sampling step ${t+1}$, the output can be expressed as: 
\begin{equation}
\begin{split}
F_{t+1}(x_{t+1})+x_{t+1},
\end{split}
\end{equation}
where $F(\cdot)$ represents the calculation content of the residual branch. 
$x$ stands for identity branch.
When we employ the feature inheritance strategy shown in Fig.~\ref{07_FeatureInheritance} (b) in the next step $t$, the output can be denoted as: 
\begin{equation}
\begin{split}
F_{t+1}(x_{t+1})+x_{t}.
\end{split}
\end{equation}

The feature inheritance strategy adopts the residual branch calculation result $F_{t+1}(x_{t+1})$ of the previous step $t+1$ to replace the calculation result $F_{t}(x_{t})$ of the current step $t$.
Since $F_{t+1}(x_{t+1})$ is calculated in the previous $t+1$ step, the time required to calculate $F_{t}(x_{t})$ can be saved in this step.

Fig.~\ref{07_FeatureInheritance} (c) and (d) depict the ResNet unit (ResUnit) and Attention unit (AttnUnit) with residual structure, which are the two main components of the UNet. 
Typically, a ResUnit and an AttnUnit constitute a layer, multiple layers form a block, and multiple blocks then constitute a complete standard UNet of SDM.
Building upon the concept of feature inheritance, we can implement it at different levels: block-level, layer-level, unit-level.
Besides, concurrent feature inheritance can also be implemented, so the proposed method has a strong generalization at the structural level.

\subsection{Layer-level Feature Inheritance} 

\begin{figure}[t]
\centering
\includegraphics[width=1\columnwidth]{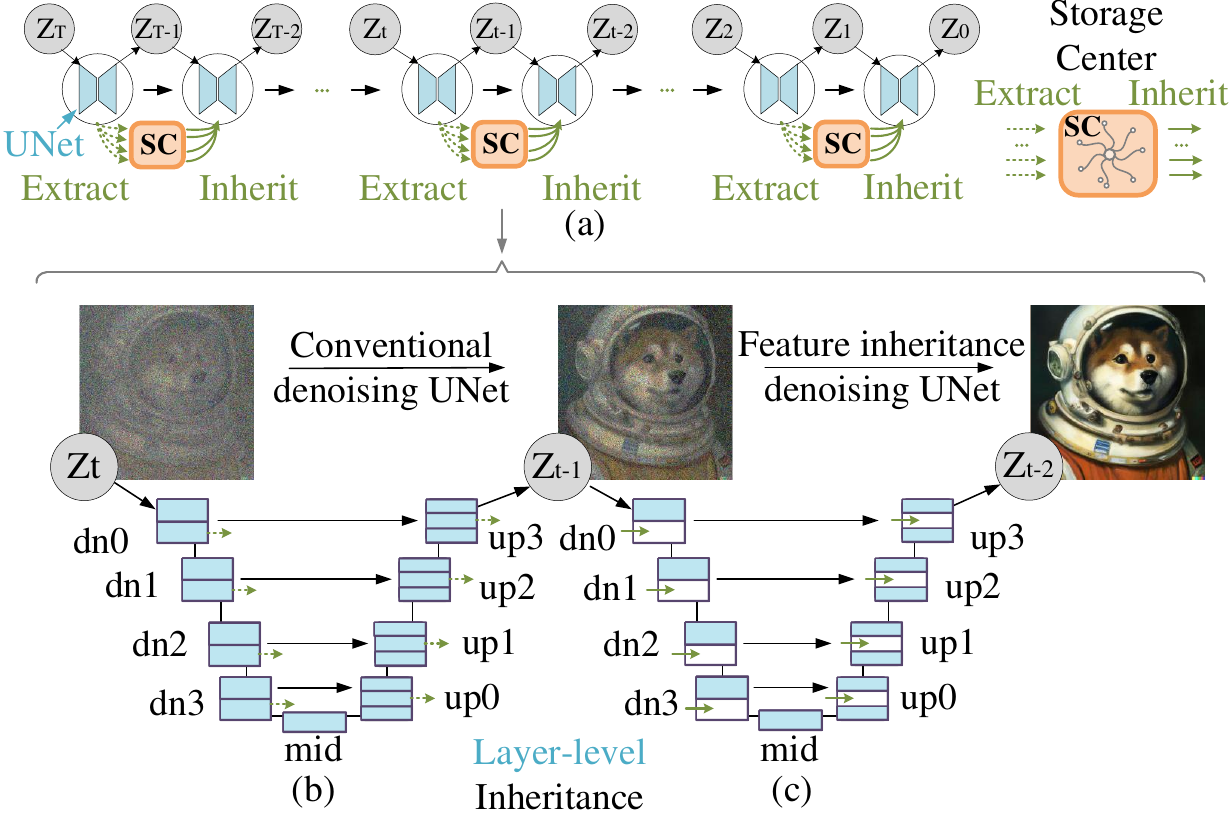}
\caption{(a) Feature inheritance process with period 2. The features of the UNet extracted in step $\text{T}$ are stored in the storage center (SC) and passed to the UNet of step $\text{T-1}$ for feature inheritance. (b) is a conventional UNet denoising process. (c) is a UNet denosing process of the layer-level feature inheritance.}
\label{08_LayerLevel}
\end{figure}

We first leverage the feature inheritance strategy to achieve layer-level computation skip, corresponding to the layers distilled away in Sec.~\ref{section_training}. 
Specifically, we skip the second layer in down blocks and the middle layer in up blocks, as depicted in Fig.~\ref{08_LayerLevel} (c). 
Fig.~\ref{08_LayerLevel} (a) shows the simplest interval inheritance method with a period of 2 steps, performing a layer-level inheritance UNet (c) inference after a conventional UNet inference (b), and then repeating the above process. 
We extract the features (dotted green arrow) of the residual branch from the ResUnit and AttnUnit in the key layer from the previous step and store these features in a storage center.
Subsequently, these stored features are inherited (solid green arrow) to the skip layer (white block in (c)) in the current step to accelerate.

It is noteworthy that although the features of the residual branch are inherited from the previous step, the existence of the identity branch enables the calculation results of the previous layer to flow smoothly through the entire UNet.
(c) demonstrates standard layer-level inheritance and we investigate further local layer skipping structures in subsequent experiments. 

\begin{figure}[t]
\centering
\includegraphics[width=0.97\columnwidth]{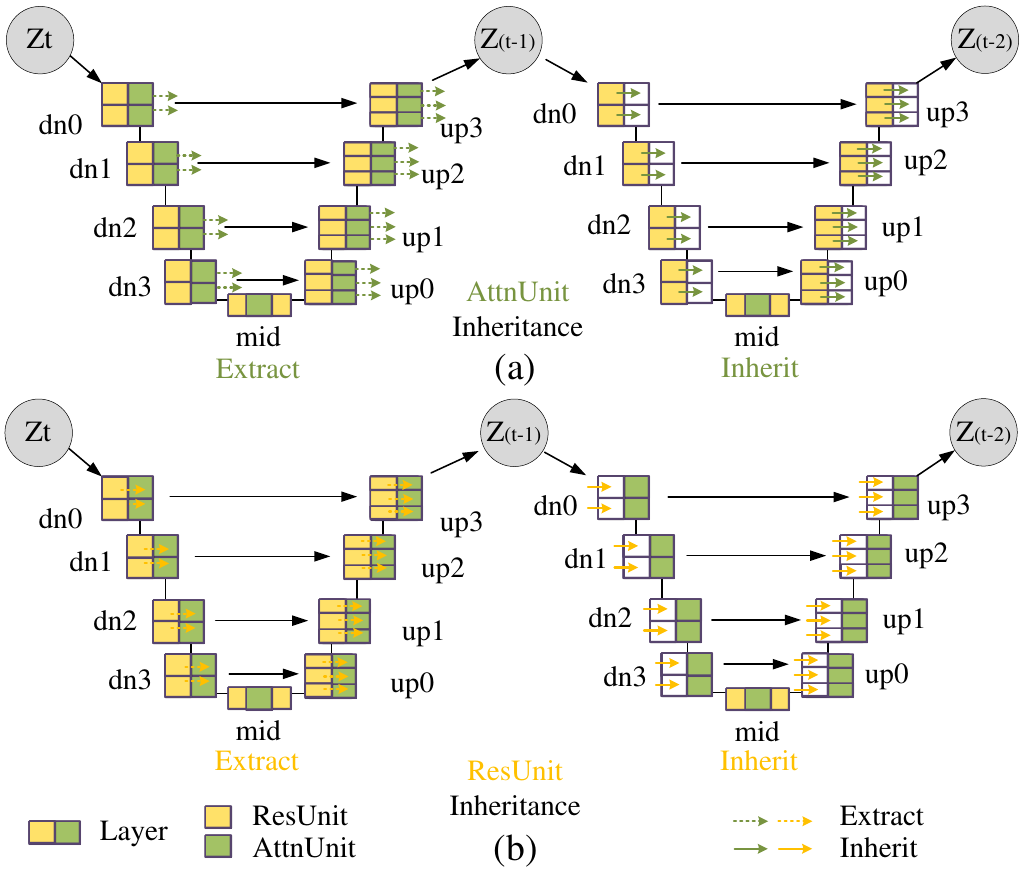}
\caption{Unit-level feature inheritance. (a) and (b) represent AttnUnit feature inheritance and ResUnit feature inheritance, respectively. Each layer is split into the ResUnit (\textcolor[rgb]{1, 0.7, 0}{yellow} square) and the AttnUnit (\textcolor[rgb]{0,0.5,0}{green} square). Dotted arrows depict feature extraction, while solid arrows represent feature inheritance.
}
\label{09_UnitLevel}
\end{figure}

\subsection{Unit-level Feature Inheritance}

Layer-level feature inheritance involves inheriting features for both AttnUnit and ResUnit of a certain layer, whereas unit-level feature inheritance entails inheriting features for either AttnUnit or ResUnit.
Fig.~\ref{09_UnitLevel} (a) and (b) illustrate the feature inheritance for AttnUnit and ResUnit, respectively. 
This section investigates the role of AttnUnit and ResUnit in the SDM denoising process.  
Similar to layer-level feature inheritance, the sampling mode with a period of 2 is adopted in the unit-level feature inheritance, and features are inherited only for units in down and up blocks. 

\begin{figure}[t]
\centering
\includegraphics[width=0.85\columnwidth]{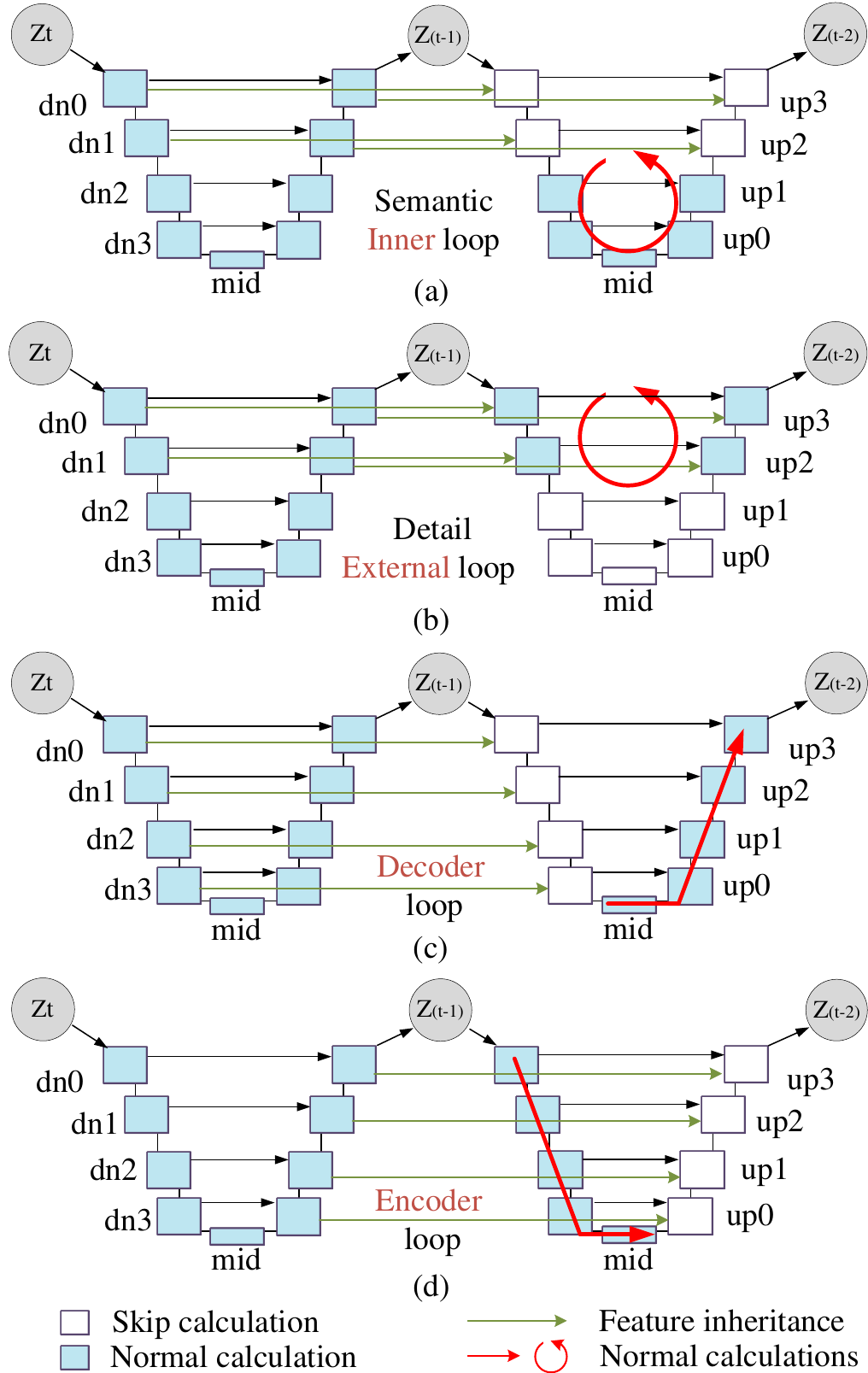}
\caption{Block-level feature inheritance. (a) Inner loop. (b) External loop. (c) Decoder loop. (d) Encoder loop. The green arrows indicate the location of the feature extraction and inheritance. The red arrow indicates the area for conventional calculations.
} 
\label{10_BlockLevel}
\end{figure}

\begin{figure}[t]
\centering
\includegraphics[width=1\columnwidth]{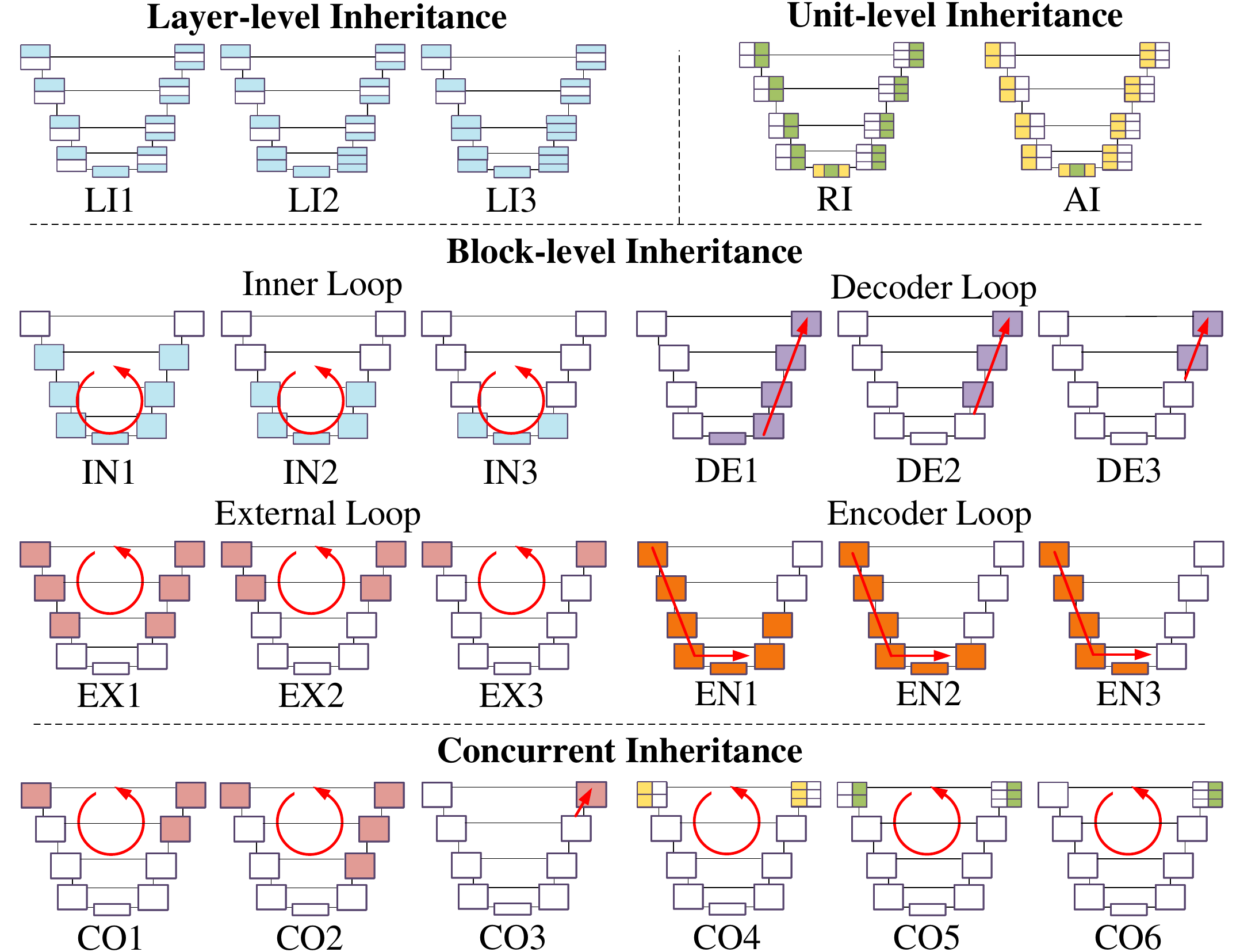}
\caption{Representative feature inheritance structure. Layer-level inheritance variants (LI1-3). Unit-level inheritance includes ResUnit inheritance (RI) and AttnUnit inheritance (AI). Inner loop variants (IN1-3). Decoder loop variants (DE1-3). External loop variants (EX1-3). Encoder loop variants (EN1-3). Block-level concurrent feature inheritance (CO1-3). Unit-level concurrent feature inheritance (CO4-5).
}
\label{11_IEDE}
\end{figure}

\subsection{Block-level Feature Inheritance}  
After designing layer-level and unit-level inheritance, our next objective is to extend the feature inheritance strategy at the block level, aiming to elucidate the contributions of shallow layers, deep layers, encoder, and decoder in the generation process.

% semantic inner loop 
In general, the deep layers of UNet are pivotal for semantic information editing. Thus, we introduce the semantic \textit{inner loop} pattern, as depicted in Fig.~\ref{10_BlockLevel} (a). 
Here, the UNet conducts standard denoising in the deep layers, while feature inheritance realizes shallow computation skipping, amplifying the semantic information iteration within the network.
% detail external loop
On the contrary, the shallow layers of the UNet play a crucial role in detail editing. Hence, we propose the detail \textit{external loop} pattern (Fig.~\ref{10_BlockLevel} (b)) allowing the UNet to process the shallow part while skipping the deep part.

% encoder loop
Furthermore, the encoder of UNet is tasked with understanding input information, while the decoder is responsible for expressing information based on the understood input.
In light of this, we introduce a \textit{decoder loop} pattern in Fig.~\ref{10_BlockLevel} (c) to bolster decoder representation while diminishing encoder understanding through feature inheritance.
% decoder loop
Conversely, in Fig.~\ref{10_BlockLevel} (d) we present the \textit{encoder loop} to reinforce input understanding while weakening decoder representation.
The above experimental design echoes the basic understanding of UNet in Fig.~\ref{02_SD_Plaincut} (a), verifying the role of UNet components in the generation process.

\subsection{Concurrent Feature Inheritance} 
Building upon the preceding research, this section introduces additional variant structures of the feature inheritance.
For brevity, Fig.~\ref{11_IEDE} showcases variants of the layer-level inheritance (LI1, LI2, LI3), external loop (EX1, EX2, EX3), inner loop (IN1, IN2, IN3), decoder loop (DE1, DE2, DE3), and encoder loop (EN1, EN2, EN3). 
Fig.~\ref{11_IEDE} focuses solely on the UNet of the feature inheritance part, omitting the feature extraction part.
White layers, blocks, and unit parts signify local feature inheritance operations.

Although numerous variations are explored, for the sake of brevity, 
the lower part of Fig.~\ref{11_IEDE} showcases the representative concurrent feature inheritance structures (CO1, CO2, CO3, CO4, CO5, CO6). 
Experiments highlight the effectiveness of decoder and external loops, prompting the addition of block-level concurrent feature inheritance (CO1-3).
Additionally, unit-level concurrent feature inheritance strategies (CO4-6) are explored further.
CO4 adopts an external loop structure, engaging solely the ResUnit part in calculations.
In CO5, only the AttnUnit part participates in calculations, while CO6 further confines AttnUnit calculations to the up3 block.
According to the experiment outcomes, it is observed that enhancing the shallow AttnUnit part significantly improves the FID metrics.

\subsection{Sampling Mode Design}
\label{SamplingMode}

The primary goal of feature inheritance is to maintain or optimize performance while improving the computational efficiency of the network.
Thus, for the same feature inheritance structure, incorporating inheritance operations into more denoising steps will greatly improve network speed.

\begin{figure}[htb]
\centering
\includegraphics[width=1\columnwidth]{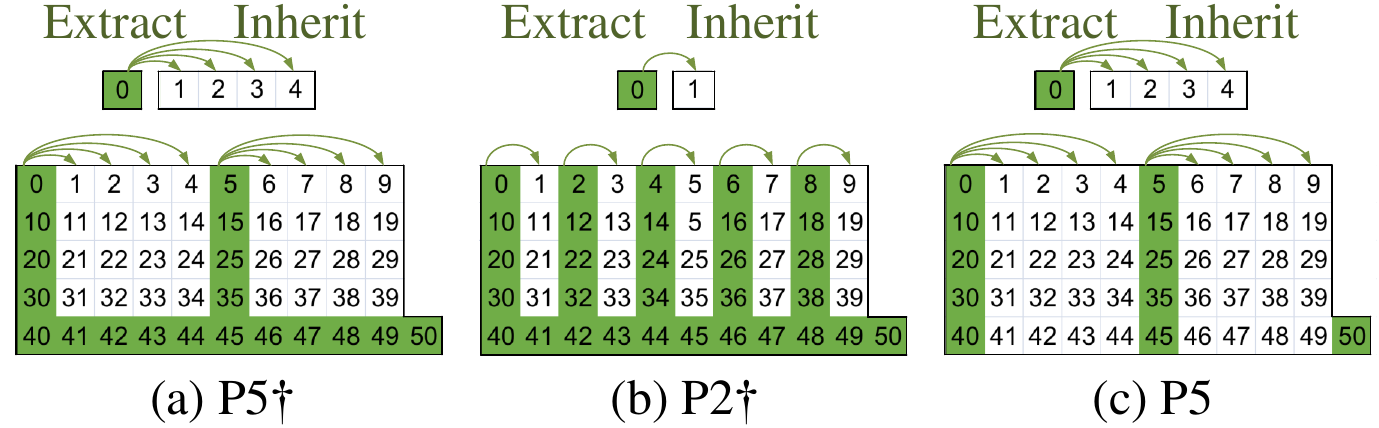}
\caption{Sampling mode design. (a) Sampling mode $P5\dag$. (b) Sampling mode $P2\dag$. (c) Sampling mode $P5$. $P5$ indicates that the feature inheritance period is 5. $\dag$ indicates that the last 10 steps do not use feature inheritance. The \textcolor[rgb]{0,0.4,0}{green} square denotes the conventional denoising step with extraction operation, and the white square represents the step performing inheritance operation.}
\label{12_mode}
\end{figure}

In this section, we delve into the sampling mode of feature inheritance.
Feature inheritance includes extraction operation and inheritance operation.
As depicted in Fig.~\ref{12_mode} (c), the $t$ denoising step (green part) executes the extraction operation, while the inheritance operation (white part) occurs in steps $t+1$, $t+2$, $t+3$, and $t+4$ steps, followed by repetition.
The entire sampling process consists of 50 steps, defining the above mode as $P5$ (feature inheritance period of 5).
Informed by the findings of Sec.~\ref{multi-unet}, where employing a small network in the early denoising stage and a complete UNet in the later stage often yields superb generation results, we introduce the specially designed $P5\dag$ sampling mode (Fig.~\ref{12_mode} (a)). 
Here, the last 10 steps perform conventional denoising operations, while periodic feature inheritance is employed in the early stage, akin to $P5$.
Similarly, as illustrated in Fig.~\ref{12_mode} (b), we devise the $P2\dag$ sampling mode with a period of 2.

\section{Experiments}

\subsection{Datasets and Metrics}
\textit{1) Model Assembly Strategy:}

Following \cite{2023-ICMLW-bksdm}, 0.22M image-text pairs from LAION Aesthetics V2 (L-Aes) 6.5+ ~\cite{2022-Laion-Aes, 2022-NIPSW-Laion-5b} are adopted as the training dataset for step 1 and step 2 of model assembly.
Additionally, consistent with approaches~\cite{2021-ICML-zeroT2I, 2022-arXiv-Hierarchical, 2022-NIPS-Photorealistic}, we utilize 30K prompts from the MS-COCO validation split~\cite{2014-ECCV-MSCOCO} to generate $512\times512$ images.
Subsequently, we downsample these images to $256\times256$ for comparison with the entire MS-COCO validation set.
%---bksdm
% primarily use 0.22M image-text pairs from LAION-Aesthetics V2 (L-Aes) 6.5+ 
% Following the popular protocol [ 49 , 55 , 60 ], we use 30K prompts from the MS-COCO validation split [ 32 ], downsample the 512×512 generated  images to 256×256, and compare them with the entire validation set.

In terms of evaluation, Fréchet Inception Distance (FID)~\cite{2017-NIPS-FID} and Inception Score (IS)~\cite{2016-NIPS-IS} are employed for visual quality assessment.
CLIP score~\cite{2021-EMNLP-CLIP, 2021-ICML-CLIP2} with CLIP-ViT-g/14 model is used to evaluate the correspondence between text and generated image.

\textit{2) Feature Inheritance Strategy:}

Feature inheritance is a tuning-free method, thus no training dataset is required. 
The validation set (MS-COCO~\cite{2014-ECCV-MSCOCO}) and evaluation metrics (FID~\cite{2017-NIPS-FID}, IS~\cite{2016-NIPS-IS}, CLIP~\cite{2021-EMNLP-CLIP, 2021-ICML-CLIP2}) are consistent with the model assembly strategy.

\subsection{Implementation Details}  
\textit{1) Model Assembly Strategy:}

% diffusers
We construct the proposed models based on the Diffusers~\footnote{\url{https://github.com/huggingface/diffusers}}.
% SD version
During the distillation process, the teacher model utilizes the SDM Runway 1.5 version (SD1.5).
The student models (both compressed and reconstructed models) are adapted from SD1.5.
% resolution
The latent resolution is set to the default $64\times64$, resulting in $512\times512$ images.
% VAE
Both the encoders and decoders of VAE perform $8\times$ downsampling and upsampling.
% classifier-free guidance scale
Additionally, the classifier-free guidance scale~\cite{2022-NIPSW-ClassifierFree, 2022-NIPS-PhotorealisticT2I} is maintained at the default value of $7.5$.
% PNDM
For sampling, we utilize the Diffusers default PNDMScheduler for all experiments.
% GPU train
During training, we leverage 8 NVIDIA V100 GPUs. %35G
% batch size 
The batch size is set to $512$. % $$512$.
%The total batch size across the 8 GPUs is set to 512 (8$\times$64).
% step
We conduct 2$\times$25K iterations during training.
The fine-tuning of the first and second steps requires 25K iterations each.
% GPU inference
A single NVIDIA GeForce RTX 2080Ti GPU is used for image generation.
% 25step
For computational efficiency and ease of comparison, we consistently employ 25 denoising steps of UNet during the inference phase in our experiments.

\textit{2) Feature Inheritance Strategy:}

The feature inheritance strategy adopts the same setup as the model assembly strategy during the inference phase.
The distinction lies in the 50 denoising steps of UNet in feature inheritance.
%We have made adjustments to the codes in Diffusers~\footnote{\url{https://github.com/huggingface/diffusers}}. 
%Our main retraining is performed using 32 NVIDIA V100 35GB GPUs.
%For computational efficiency, we consistently opt for 30 denoising steps of the UNet during the inference phase, unless otherwise specified.
%The classifier-free guidance scale is set to the default value of $8.0$. Additionally, the latent resolution is set to $512\times512$.

\subsection{Experimental Analysis} 

% i2i 
\textit{1) Model Assembly Strategy:}

\textbf{Ablation studies: }
Tab.~\ref{Tab-M123} shows the evaluation results of the three reconstructed models M1, M2, and M3 in Fig.~\ref{04_Incubation_Micro} of Sec.~\ref{model compression}.
Take the M1 (Base$_\text{dn0+up3}$+SD$_\text{dn123+up012}$) model as an example. Structurally, it assembles a portion (dn0, up3) of Base-UNet and a portion (dn1-3, up0-2) of SD-UNet.
The shallow layer of M1 comprises a more compact Base-UNet, making it easier to learn via distillation due to its fewer parameters and responsibility for detailed expression in the generation process. 
The training dataset is a relatively high-quality small dataset introduced by \cite{2023-ICMLW-bksdm}, making it suitable for shallow optimization tuning.
Subsequently, we alter the structure of the reconstructed models to obtain the M2 and M3. 
Across M1 to M3, the compact Base-UNet occupies an increasing proportion of the overall structure.
Through experiments, we observe that the M2 structure yields the best results (FID $11.840$), surprisingly surpassing the original standard UNet (FID $12.832$).
As per the previous experimental analysis of Fig.~\ref{01_SD_Analysis}, the shallow layer accounts for a significant amount of calculation time. 
Therefore, despite M2 having more parameters than the Base model, there is minimal difference in inference speed.
Both M2 (1.643s) and Base (1.529s) are faster than the original SDM (2.128s) in 25-step generation tests.
In summary, M2 exhibits superior experimental results, with both speed and performance far outperforming those of SDM.

For the model assembly process, we freeze the deep part of the reconstructed model (M1-3) to preserve the stable semantic information of the deep layers.
Given that the original SDM is trained on a large dataset of 1.04 B, retraining on the 0.22M dataset from \cite{2023-ICMLW-bksdm} may have a negative effect on the deep parameters.
To validate the advantages of freezing tuning, non-freezing experiments are conducted on models M1$\ddag$, M2$\ddag$, and M3$\ddag$ for comparison, wherein the deep part of the student model remains unfrozen.
The experiments demonstrate that the frozen tuning method performs better.

\begin{table}[t]
\centering
\caption{Comparison experiments of multiple reconstructed models.
SD-UNet refers to the standard UNet of Runway 1.5. Base-UNet represents the simplified UNet obtained through distillation of the BK-SDM method.
M1, M2, and M3 denote three different UNet assembly structures.
For example, in M1, Base$_\text{dn0+up3}$+SD$_\text{dn123+up012}$ indicates that the current reconstructed UNet adopts down0 and up3 blocks of the Base-UNet, with the remaining parts adopting down1, down2, down3, up0, up1, and up2 blocks of SD-UNet.
* signifies that the SD-UNet part of the reconstructed UNet is frozen during the second distillation process. The sampling step is 25. \textcolor{red}{Red} marks indicate best performance.
}
\setlength{\tabcolsep}{1mm}{  
\begin{tabular}{lccc}
	\hline
	\multicolumn{1}{c}{\multirow{2}*{Method}} & \multicolumn{3}{c}{Metrics} \\
	\cline{2-4}          & FID$\downarrow$   & IS$\uparrow$ & CLIP$\uparrow$ \\
	\hline
	Standard SD-UNet & 12.832 & 36.653 & 0.297 \\
	\hline
	M1: Base$_\text{dn0+up3}$+SD*$_\text{dn123+up012}$ & 13.049 & 37.848 & 0.298 \\
	M2: Base$_\text{dn01+up23}$+SD*$_\text{dn23+up01}$ & \textcolor{red}{11.840} & 36.560 & 0.296   \\ 
	M3: Base$_\text{dn012+up123}$+SD*$_\text{dn3+up0}$ & 20.037 & 22.338 & 0.249 \\
	\hline % $^{w}_{\beta}$
	M1$\ddagger$: Base$_\text{dn0+up3}$+SD$_\text{dn123+up012}$ & 13.377 & \textcolor{red}{37.957}  & 0.298 \\
	M2$\ddagger$: Base$_\text{dn01+up23}$+SD$_\text{dn23+up01}$ & 13.222 & 37.726 & \textcolor{red}{0.299}   \\ 
	M3$\ddagger$: Base$_\text{dn012+up123}$+SD$_\text{dn3+up0}$ & 14.546 & 33.718 & 0.291   \\ 
	\hline
	Base-UNet  & 14.426 & 33.403 & 0.288 \\
	\hline
\end{tabular}%
}
\label{Tab-M123}%
\end{table}  %14.610	31.440	.288  
% 14.546 33.718 0.291

\textbf{ME-CondConv study:}
We seek to enhance the capacity of the compact models (Base, Small, Tiny) proposed by BK-SDM using ME-CondConv without increasing computational latency.
Tab.~\ref{Tab-condconv} illustrates the replacement of $3\times3$ convolution in all compact models with ME-CondConv.
Notably, when employing 2 experts, Small with ME-CondConv and Tiny with ME-CondConv exhibit significant score increases, while Base with ME-CondConv shows a decline in performance.
This disparity can be attributed to the extensive pruning of deep modules in Small and Tiny models, necessitating capacity augmentation through additional experts, unlike the Base model, which does not face such constraints.

However, the random initialization of experts in the Base model may adversely affect its performance, particularly with limited training data (0.22M).
Furthermore, increasing the number of experts does not necessarily lead to further performance enhancements in Small and Tiny models. 
This is because additional randomly initialized experts may dilute the influence of the original convolution kernel $W_{n}$ (Fig.~\ref{05_ME-CondConv} (d)), thereby impairing the network's generation capabilities.
Given the subpar performance of Base with ME-CondConv, we opt against integrating ME-CondConv into the reconstructed models M1, M2, and M3, which are built upon the Base model.

\begin{table}[htbp]
\centering
\caption{Comparative experiment with or without ME-CondConv. 
The number of experts in ME-CondConv is set to 2.
The weight of the first expert is inherited from the convolution weight of the teacher UNet (Runway 1.5 version), while the weight of the other expert is randomly initialized. Base, Small, and Tiny are the models provided by BK-SDM~\cite{2023-ICMLW-bksdm}. The sampling step is 25. \textcolor{red}{Red} marks indicate the best results. % The red marks indicate that ME-CondConv has a positive effect.
}
\setlength{\tabcolsep}{1mm}{  
\begin{tabular}{lccc}
	\hline
	\multicolumn{1}{c}{\multirow{2}*{Method}} & \multicolumn{3}{c}{Metrics} \\
	\cline{2-4}             & FID$\downarrow$   & IS$\uparrow$ & CLIP$\uparrow$ \\
	\hline
	Base-UNet                     & \textcolor{red}{14.426}	& 33.403 & \textcolor{red}{0.288} \\
	\rowcolor{gray!30}
	Base with ME-CondConv  & 14.963    & \textcolor{red}{33.522} & \textcolor{red}{0.288}   \\
	\hline
	Small-UNet                  & 17.120    & 31.271 & 0.268 \\
	\rowcolor{gray!30}
	Small with ME-CondConv 
	& \textcolor{red}{16.056} & \textcolor{red}{31.740} & \textcolor{red}{0.274} \\
	\hline
	Tiny-UNet                     & 18.670    & 27.730 & 0.257 \\
	\rowcolor{gray!30}
	Tiny with ME-CondConv  
	& \textcolor{red}{17.470} & \textcolor{red}{28.720} & \textcolor{red}{0.263}   \\  
	\hline
\end{tabular}%
}
\label{Tab-condconv}%
\end{table}  %14.610	31.440	.288

\begin{table}[htbp]
\centering
\caption{Experiments of the multi-UNet switching method. In Strategy 1 (S1), the UNet of the Base model is adopted for the first 10 steps, followed by the UNet of the standard SDM for the next 15 steps. S2 and S3 is the opposite. \textcolor{red}{Red} marks indicate the best results in S1, S2, and S3.
}  % S3 employs the UNet of the standard SDM for the initial 15 steps and the UNet of the Base model for the subsequent 10 steps. S2 adopts the standard UNet for the first 15 steps and the Base-UNet for the subsequent 10 steps.
\setlength{\tabcolsep}{1mm}{  
\begin{tabular}{lccc}
	\hline
	\multicolumn{1}{c}{\multirow{2}*{Method}} & \multicolumn{3}{c}{Metrics} \\
	\cline{2-4}             & FID$\downarrow$   & IS$\uparrow$ & CLIP$\uparrow$ \\
	\hline
	Standard SD-UNet & 12.832 & 36.653 & 0.297    \\
	\hline
	S1: Base$_\text{10 step}$ + SD$_\text{15 step}$ 
	& \textcolor{red}{12.900} & 35.651 & \textcolor{red}{0.297} \\ 
	S2: SD$_\text{15 step}$ + Base$_\text{10 step}$ 
	& 13.529 & 34.560 & 0.294 \\ 
	S3: SD$_\text{10 step}$ + Base$_\text{15 step}$                    
	& 13.548 & \textcolor{red}{36.720} & 0.295 \\ 
	\hline
	Base-UNet & 14.426 & 33.403 & 0.288\\
	\hline
\end{tabular}%
}
\label{Tab-strategy}%
\end{table}  %14.610	31.440	.288

\textbf{Multi-UNet switching study:} 
Tab.~\ref{Tab-strategy} presents the experimental outcomes of multi-UNet switching processing.
In the T2I task, the initial input is Gaussian noise, and the generated image is acquired through the cyclic process of UNet de-noising.
Initially, the input is a chaotic, low-information, noise-like image, making a compact network (Base) efficient for processing.
However, as the process advances, the input image begins to exhibit a preliminary semantic structure, necessitating a high-performance UNet (SDM) to understand and optimize it comprehensively.
In Tab.~\ref{Tab-strategy}, Strategy 1 (S1) employs a compressed model (Base) for the initial processing of 10 steps in the early stage, followed by SDM for optimization over 15 steps in the later stage.
Conversely, Strategies 2 and 3 (S2, S3) have the opposite setup for comparative experiments.
The experimental comparison reveals that the S1 strategy exhibits commendable performance, being both efficient and rapid.
This conclusion is also extended to tuning-free feature inheritance strategies.

% \resizebox{\textwidth}{!}{
% \setlength{\tabcolsep}{1.5mm}{
% [h!t]% 
\begin{table*}[htbp]
	\centering
	\caption{Comparison of results from multiple methods on zero-shot MS-COCO $256\times256$ 30K.
		The training resources part includes the size of image-text pairs, batch size, iterations, and A100 days.
		$\dag$: Evaluated with released checkpoints. 
		$\ddag$: Total parameters for T2I synthesis. 
		*: Estimated based on public information.
		DF and AR: diffusion and autoregressive models. 
		$\downarrow$ and $\uparrow$: lower and higher values are better.
		It is worth noting that A100 is used as the standard to facilitate the unified comparison of computing resources. The sampling step is 25. Part of the data comes from~\cite{2023-ICMLW-bksdm}.} % In the generation score part, the proposed M2 model in the generation score part actually uses 8 V100s for equivalent training.
	\resizebox{\textwidth}{!}{
		\begin{tabular}{lcc|ccc|ccc}
			\hline
			\multicolumn{3}{c|}{Model} & \multicolumn{3}{c|}{Generation Score} & \multicolumn{3}{c}{Training Resource} \\
			\hline
			\multicolumn{1}{l}{Name} & Type  & \# Param$\ddag$ & FID$\downarrow$   & IS$\uparrow$    & CLIP$\uparrow$  & DataSize & (Batch,\#Iter) & A100 Days \\
			\hline
			SDM-v1.4$\dag$ 
			& DF    & 1.04B & 13.05 & 36.76 & 0.2958 & $>$2000M* & (2048, 1171K)  & 6250 \\
			Small Stable Diffusion$\dag$ & DF    & 0.76B & 12.76 & 32.33 & 0.2851 & 229M  & (128, 1100K)  & - \\
			BK-SDM-Base  & DF    & 0.76B & 14.43 & 33.40 & 0.2880 & 0.22M & (256, 50K) & 13 \\
			
			BK-SDM-Small  & DF    & 0.66B 
			& 17.12 & 31.27 & 0.2680
			& 0.22M & (256, 50K) & 13 \\
			\rowcolor{gray!30}
			BK-SDM-Small with ME-CondConv$\dag$ & DF    & 0.89B 
			& 16.06   &  31.74 & 0.2735 
			& 0.22M & (256, 50K) & 13 \\
			BK-SDM-Tiny  & DF    & 0.50B 
			& 18.67 & 27.73 & 0.2570 
			& 0.22M & (256, 50K) & 13 \\
			\rowcolor{gray!30}
			BK-SDM-Tiny with ME-CondConv$\dag$ & DF    & 0.62B 
			& 17.47 & 28.72  & 0.2630 
			& 0.22M & (256, 50K) & 13 \\
			
			\hline
			SDM-v2.1-base$\dag$ 
			& DF    & 1.26B & 13.93 & 35.93 & 0.3075 & $>$2000M* & (2048, 1620K)  & 8334 \\
			BK-SDM-v2-Base$\dag$ & DF    & 0.98B & 15.85 & 31.7  & 0.2868 & 0.22M & (128, 50K) & 4 \\
			BK-SDM-v2-Small$\dag$ & DF    & 0.88B & 16.61 & 31.73 & 0.2901 & 0.22M & (128, 50K) & 4 \\
			BK-SDM-v2-Tiny$\dag$ & DF    & 0.72B & 15.68 & 31.64 & 0.2897 & 0.22M & (128, 50K) & 4 \\
			\hline
			DALL·E & AR    & 12B   & 27.5  & 17.9  & -     & 250M  & (1024, 430K) & x \\
			CogView & AR    & 4B    & 27.1  & 18.2  & -     & 30M   & (6144, 144K) & - \\
			CogView2 & AR    & 6B    & 24    & 22.4  & -     & 30M   & (4096, 300K   & x \\
			Make-A-Scene &  AR   & 4B    & 11.84 & -     & -     & 35M   &  (1024, 170K)   & - \\
			LAFITE & GAN   & 0.23B & 26.94 & 26.02 & -     & 3M    & -     & - \\
			GALIP (CC12M)$\dag$  & GAN   & 0.32B & 13.86 & 25.16 & 0.2817 & 12M   & -     & - \\
			GigaGAN  & GAN   & 1.1B  & 9.09  & -     & -     & $>$100M* & (512, 1350K)  & 4783 \\
			GLIDE  &  DF   & 3.5B  & 12.24 & -     & -     & 250M  & (2048, 2500K) & - \\
			LDM-KL-8-G   & DF    & 1.45B & 12.63 & 30.29 & -     & 400M  &  (680, 390K)   & - \\
			DALL·E-2  & DF    & 5.2B  & 10.39 & -     & -     & 250M  &  (4096, 3400K)   & - \\
			SnapFusion  & DF    & 0.99B & $\sim$13.6  & -     & $\sim$0.295 & $>$100M* & (2048, -) &  $>$128* \\
			Würstchen-v2$\dag$ & DF    & 3.1B  & 22.4  & 32.87 & 0.2676 & 1700M &  (1536, 1725K)  & 1484 \\
			\hline
			\rowcolor{gray!30}
			M2$\dag$: Base$_\text{dn01+up23}$+SD*$_\text{dn23+up01}$          & DF    & 0.98B  
			& 11.84   & 36.56     & 0.2958 
			& 0.22M &  (512, 2$\times$25K)  & $<$26 \\
			\hline
		\end{tabular}
	}
	\label{PI-SOTA}%
\end{table*}%
%16.056   &  31.740 & 0.274
%17.470 & 28.720  & 0.263

\begin{figure*}[htbp]
	\centering
	\includegraphics[width=2\columnwidth]{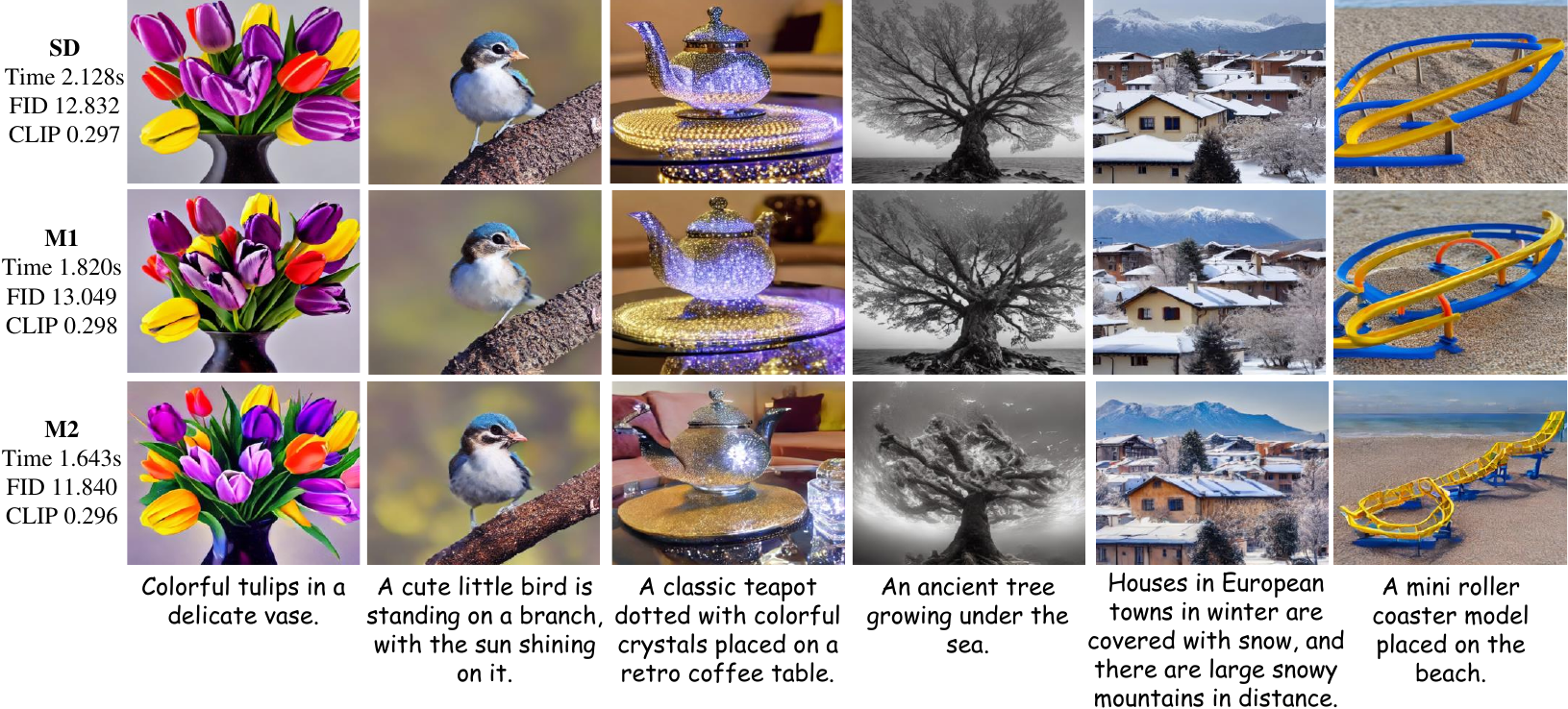}
	\caption{Comparison of the generation results between models (M1, M2) obtained by model assembly strategy and SDM. The sampling step is 25.
	}
	\label{13_Distillation}
\end{figure*} 

%%%

%The comparison between the results of feature inheritance strategies (EX3, CO3, CO5, CO6) and the results of SDM without feature inheritance

% Table generated by Excel2LaTeX from sheet 
\begin{table*}[t] % 
	\centering
	\caption{Qualitative comparison of feature inheritance strategies, including layer-level (LI1-3), unit-level (AI and RI), internal loop (IN1-3), external loop (EX1-3), encoder loop (EN1-3), decoder loop (DE1-3), and concurrent feature inheritance (CO1-6). Different sampling strategies ($P5\dag$, $P2\dag$, and $P5$) are evaluated, where $P5$ involves one complete UNet calculation followed by four feature inheritance UNet calculations every five denoising steps, and $P2$ alternates between full UNet and feature inheritance calculations. $\dag$ denotes that the last 10 denoising steps do not utilize feature inheritance. Scores less than 11 in FID are highlighted in \textcolor{red}{red}.
	}
	\begin{tabular}{l|rrr|rrr|rrr}
		\hline
		Metrics & \multicolumn{1}{c}{FID$\downarrow$} & \multicolumn{1}{c}{IS$\uparrow$} & \multicolumn{1}{c|}{CLIP$\uparrow$} & \multicolumn{1}{c}{FID$\downarrow$} & \multicolumn{1}{c}{IS$\uparrow$} & \multicolumn{1}{c|}{CLIP$\uparrow$} & \multicolumn{1}{c}{FID$\downarrow$} & \multicolumn{1}{c}{IS$\uparrow$} & \multicolumn{1}{c}{CLIP$\uparrow$} \\
		\hline
		\multicolumn{10}{c}{Feature Inheritance (50step)} \\
		\hline
		Strategy & \multicolumn{3}{c|}{ Sample mode1: $P5\dag$} & \multicolumn{3}{c|}{ Sampling mode2: $P2\dag$} & \multicolumn{3}{c}{ Sample mode3: $P5$} \\
		\hline
		LI1  (layer-level)  & \multicolumn{1}{c}{\textcolor[rgb]{ 1,  0,  0}{10.861}} & \multicolumn{1}{c}{37.026} & \multicolumn{1}{c|}{0.297} & \multicolumn{1}{c}{12.062} & \multicolumn{1}{c}{37.527} & \multicolumn{1}{c|}{0.299} & \multicolumn{1}{c}{\textcolor[rgb]{ 1,  0,  0}{10.939}} & \multicolumn{1}{c}{36.757} & \multicolumn{1}{c}{0.298} \\
		LI2  (layer-level)  & \multicolumn{1}{c}{\textcolor[rgb]{ 1,  0,  0}{10.964}} & \multicolumn{1}{c}{36.879} & \multicolumn{1}{c|}{0.299} & \multicolumn{1}{c}{12.455} & \multicolumn{1}{c}{37.506} & \multicolumn{1}{c|}{0.299} & \multicolumn{1}{c}{11.032} & \multicolumn{1}{c}{36.820} & \multicolumn{1}{c}{0.298} \\
		LI3 (layer-level)   & \multicolumn{1}{c}{14.816} & \multicolumn{1}{c}{35.602} & \multicolumn{1}{c|}{0.297} & \multicolumn{1}{c}{12.490} & \multicolumn{1}{c}{37.304} & \multicolumn{1}{c|}{0.300} & \multicolumn{1}{c}{11.713} & \multicolumn{1}{c}{37.257} & \multicolumn{1}{c}{0.298} \\
		\hline
		AI (unit-level) & \multicolumn{1}{c}{16.194} & \multicolumn{1}{c}{34.235} & \multicolumn{1}{c|}{0.295} & \multicolumn{1}{c}{14.541} & \multicolumn{1}{c}{36.670} & \multicolumn{1}{c|}{0.299} & \multicolumn{1}{c}{15.564} & \multicolumn{1}{c}{33.862} & \multicolumn{1}{c}{0.291} \\
		RI (unit-level) & \multicolumn{1}{c}{\textcolor[rgb]{ 1,  0,  0}{10.360}} & \multicolumn{1}{c}{36.833} & \multicolumn{1}{c|}{0.298} & \multicolumn{1}{c}{11.654} & \multicolumn{1}{c}{37.800} & \multicolumn{1}{c|}{0.300} & \multicolumn{1}{c}{\textcolor[rgb]{ 1,  0,  0}{10.866}} & \multicolumn{1}{c}{36.183} & \multicolumn{1}{c}{0.297} \\
		\hline
		IN1 (block-level) & \multicolumn{1}{c}{15.259} & \multicolumn{1}{c}{34.953} & \multicolumn{1}{c|}{0.296} & \multicolumn{1}{c}{14.360} & \multicolumn{1}{c}{37.056} & \multicolumn{1}{c|}{0.298} & \multicolumn{1}{c}{15.921} & \multicolumn{1}{c}{33.965} & \multicolumn{1}{c}{0.290} \\
		IN2 (block-level) & \multicolumn{1}{c}{14.411} & \multicolumn{1}{c}{34.868} & \multicolumn{1}{c|}{0.296} & \multicolumn{1}{c}{13.985} & \multicolumn{1}{c}{37.033} & \multicolumn{1}{c|}{0.299} & \multicolumn{1}{c}{15.546} & \multicolumn{1}{c}{33.828} & \multicolumn{1}{c}{0.289} \\
		IN3 (block-level) & \multicolumn{1}{c}{14.457} & \multicolumn{1}{c}{34.559} & \multicolumn{1}{c|}{0.296} & \multicolumn{1}{c}{14.033} & \multicolumn{1}{c}{36.915} & \multicolumn{1}{c|}{0.299} & \multicolumn{1}{c}{15.625} & \multicolumn{1}{c}{33.296} & \multicolumn{1}{c}{0.289} \\
		\hline
		EX1 (block-level) & \multicolumn{1}{c}{12.988} & \multicolumn{1}{c}{36.624} & \multicolumn{1}{c|}{0.297} & \multicolumn{1}{c}{12.924} & \multicolumn{1}{c}{37.267} & \multicolumn{1}{c|}{0.299} & \multicolumn{1}{c}{12.999} & \multicolumn{1}{c}{36.618} & \multicolumn{1}{c}{0.298} \\
		EX2 (block-level) & \multicolumn{1}{c}{12.575} & \multicolumn{1}{c}{36.365} & \multicolumn{1}{c|}{0.297} & \multicolumn{1}{c}{12.589} & \multicolumn{1}{c}{37.459} & \multicolumn{1}{c|}{0.299} & \multicolumn{1}{c}{12.677} & \multicolumn{1}{c}{36.443} & \multicolumn{1}{c}{0.297} \\
		EX3 (block-level) & \multicolumn{1}{c}{11.115} & \multicolumn{1}{c}{36.487} & \multicolumn{1}{c|}{0.298} & \multicolumn{1}{c}{12.106} & \multicolumn{1}{c}{37.397} & \multicolumn{1}{c|}{0.300} & \multicolumn{1}{c}{11.342} & \multicolumn{1}{c}{36.428} & \multicolumn{1}{c}{0.298} \\% 
		\hline
		EN1 (block-level) & \multicolumn{1}{c}{ 14.916} & \multicolumn{1}{c}{ 35.396} & \multicolumn{1}{c|}{0.297} & \multicolumn{1}{c}{14.418} & \multicolumn{1}{c}{36.567} & \multicolumn{1}{c|}{0.298} & \multicolumn{1}{c}{15.757} & \multicolumn{1}{c}{34.439} & \multicolumn{1}{c}{0.291} \\
		EN2 (block-level) & \multicolumn{1}{c}{14.488} & \multicolumn{1}{c}{34.463} & \multicolumn{1}{c|}{0.296} & \multicolumn{1}{c}{14.144} & \multicolumn{1}{c}{36.723} & \multicolumn{1}{c|}{0.299} & \multicolumn{1}{c}{15.485} & \multicolumn{1}{c}{33.257} & \multicolumn{1}{c}{0.289} \\
		EN3 (block-level) & \multicolumn{1}{c}{14.512} & \multicolumn{1}{c}{34.323} & \multicolumn{1}{c|}{0.296} & \multicolumn{1}{c}{14.100} & \multicolumn{1}{c}{36.755} & \multicolumn{1}{c|}{0.299} & \multicolumn{1}{c}{15.528} & \multicolumn{1}{c}{33.135} & \multicolumn{1}{c}{0.289} \\
		\hline
		DE1 (block-level) & \multicolumn{1}{c}{11.853} & \multicolumn{1}{c}{36.629} & \multicolumn{1}{c|}{0.296} & \multicolumn{1}{c}{12.441} & \multicolumn{1}{c}{37.603} & \multicolumn{1}{c|}{0.299} & \multicolumn{1}{c}{11.884} & \multicolumn{1}{c}{36.489} & \multicolumn{1}{c}{0.296} \\
		DE2 (block-level) & \multicolumn{1}{c}{11.857} & \multicolumn{1}{c}{36.519} & \multicolumn{1}{c|}{0.296} & \multicolumn{1}{c}{12.396} & \multicolumn{1}{c}{37.584} & \multicolumn{1}{c|}{0.299} & \multicolumn{1}{c}{11.873} & \multicolumn{1}{c}{36.331} & \multicolumn{1}{c}{0.296} \\
		DE3 (block-level) & \multicolumn{1}{c}{11.771} & \multicolumn{1}{c}{36.571} & \multicolumn{1}{c|}{0.297} & \multicolumn{1}{c}{12.313} & \multicolumn{1}{c}{37.472} & \multicolumn{1}{c|}{0.299} & \multicolumn{1}{c}{11.844} & \multicolumn{1}{c}{36.472} & \multicolumn{1}{c}{0.296} \\
		\hline
		CO1 (concurrent) & \multicolumn{1}{c}{12.371} & \multicolumn{1}{c}{36.349} & \multicolumn{1}{c|}{0.296} & \multicolumn{1}{c}{12.523} & \multicolumn{1}{c}{37.539} & \multicolumn{1}{c|}{0.299} & \multicolumn{1}{c}{12.472} & \multicolumn{1}{c}{36.194} & \multicolumn{1}{c}{0.296} \\
		CO2 (concurrent) & \multicolumn{1}{c}{12.602} & \multicolumn{1}{c}{36.167} & \multicolumn{1}{c|}{0.296} & \multicolumn{1}{c}{12.635} & \multicolumn{1}{c}{37.171} & \multicolumn{1}{c|}{0.299} & \multicolumn{1}{c}{12.698} & \multicolumn{1}{c}{36.098} & \multicolumn{1}{c}{0.296} \\
		CO3 (concurrent) & \multicolumn{1}{c}{\textcolor[rgb]{ 1,  0,  0}{10.961}} & \multicolumn{1}{c}{36.581} & \multicolumn{1}{c|}{0.298} & \multicolumn{1}{c}{12.066} & \multicolumn{1}{c}{37.677} & \multicolumn{1}{c|}{0.300} & \multicolumn{1}{c}{11.136} & \multicolumn{1}{c}{36.459} & \multicolumn{1}{c}{0.298} \\
		CO4 (concurrent) & \multicolumn{1}{c}{15.591} & \multicolumn{1}{c}{33.890} & \multicolumn{1}{c|}{0.295} & \multicolumn{1}{c}{14.368} & \multicolumn{1}{c}{36.647} & \multicolumn{1}{c|}{0.299} & \multicolumn{1}{c}{15.224} & \multicolumn{1}{c}{33.750} & \multicolumn{1}{c}{0.291} \\
		CO5 (concurrent) & \multicolumn{1}{c}{\textcolor[rgb]{ 1,  0,  0}{10.405}} & \multicolumn{1}{c}{36.791} & \multicolumn{1}{c|}{0.298} & \multicolumn{1}{c}{11.659} & \multicolumn{1}{c}{37.738} & \multicolumn{1}{c|}{0.300} & \multicolumn{1}{c}{\textcolor[rgb]{ 1,  0,  0}{10.896}} & \multicolumn{1}{c}{35.930} & \multicolumn{1}{c}{0.297} \\
		CO6 (concurrent) & \multicolumn{1}{c}{\textcolor[rgb]{ 1,  0,  0}{10.406}} & \multicolumn{1}{c}{36.677} & \multicolumn{1}{c|}{0.298} & \multicolumn{1}{c}{11.670} & \multicolumn{1}{c}{37.801} & \multicolumn{1}{c|}{0.300} & \multicolumn{1}{c}{\textcolor[rgb]{ 1,  0,  0}{10.867}} & \multicolumn{1}{c}{35.993} & \multicolumn{1}{c}{0.297} \\
		\hline
		\multicolumn{10}{c}{None Feature Inheritance} \\
		\hline
		Strategy & \multicolumn{3}{c|}{Normal sampling 50step} & \multicolumn{3}{c|}{Normal sampling 25step} & \multicolumn{3}{c}{Normal sampling 10step} \\ 
		\hline
		Normal sampling & 13.177 & 37.389 & 0.298 & 12.832 & 36.653 & 0.297 & 14.963    & 29.109   & 0.276 \\
		\hline
	\end{tabular}%
	\label{FI-SOTA}%
\end{table*}%

\textbf{Quantitative and qualitative analysis:}
% speed  quantitative qualitative
Tab.~\ref{PI-SOTA} illustrates the performance of the proposed method (M2) compared to other methods.
The proposed reconstructed model M2, obtained through the model assembly strategy, exhibits clear advantages in terms of generation score compared to most methods, including BK-SDM~\cite{2023-ICMLW-bksdm}, SDM~\cite{2022-CVPR-LDM}, and SnapFusion~\cite{2023-nips-snapfusion}.
Moreover, the proposed model significantly reduces the number of parameters compared to SDM.
Despite having more parameters than the BK-SDM-Base model, M2 still shows significantly improved speed due to pruning for shallow layers of computational redundancy.
The single image generation speeds (25 steps) of SDM, BK-SDM-Base, and M2 on a 2080Ti GPU are 2.128s, 1.529s, and 1.643s, respectively. Considering both speed and performance comprehensively, the M2 model is superior.

It is worth noting that the required training resources are comparable to those of previous work~\cite{2023-ICMLW-bksdm}.
%For uniform comparison, the training time is represented using A100 in Tab.~\ref{PI-SOTA}. 
For uniform comparison, the training time is measured by A100 days in Tab.~\ref{PI-SOTA}.
%measured in Tab.~\ref{PI-SOTA} using A100 days. 
In practice, we utilized 8 V100s to train the M2 model on equivalent conditions.  % for training equivalent to M2.
In the first step of model assembly, as in~\cite{2023-ICMLW-bksdm}, the batch size is set to $512$ and the iteration to 25K.
The setup for the second step of model assembly is identical, resulting in a total of 50K iterations.
Since the second step involves partial freezing, the training time is less than 13 A100 days, with a total training time of under 26 A100 days.

% In summary, the progressive distillation strategy can obtain an efficient generation model with better performance (11.84 FID), fewer parameters (0.98B) and faster speed (3.107s) under the constraints of small data (0.22M), low computational requirements (one A100) and short training time ($<$26 A100 days).
In summary, the model assembly strategy enables the acquisition of an efficient generation model with improved performance, reduced parameters, and faster speed, all within the constraints of limited data (0.22M), low computational requirements (single A100 GPU), and short training time (less than 26 A100 days). 
Fig.~\ref{13_Distillation} compares the generated results of the models (M1, M2) obtained by the model assembly strategy with the output results of the teacher model SDM.  
Refer to \ref{appendix-b} for more visualizations, which comprehensively compare SDM, M1, M2, Base, Small, and Tiny models.
% Refer to \ref{appendix-b} for more visual results.

\textit{2) Feature Inheritance Strategy:}

\textbf{Feature inheritance structure study:}
As displayed in Tab.~\ref{FI-SOTA}, we compare different feature inheritance structures using the $P5\dag$ sampling mode.
% layer-level
\textit{Layer-level} feature inheritance LI1, which skips one layer per block, demonstrates superior performance and efficient speedup.
Structurally, when the feature inheritance operation area is reduced (LI2 and LI3 in Fig.~\ref{11_IEDE}), the performance on the FID metric deteriorates.

\begin{figure*}[t]
	\centering
	\includegraphics[width=2\columnwidth]{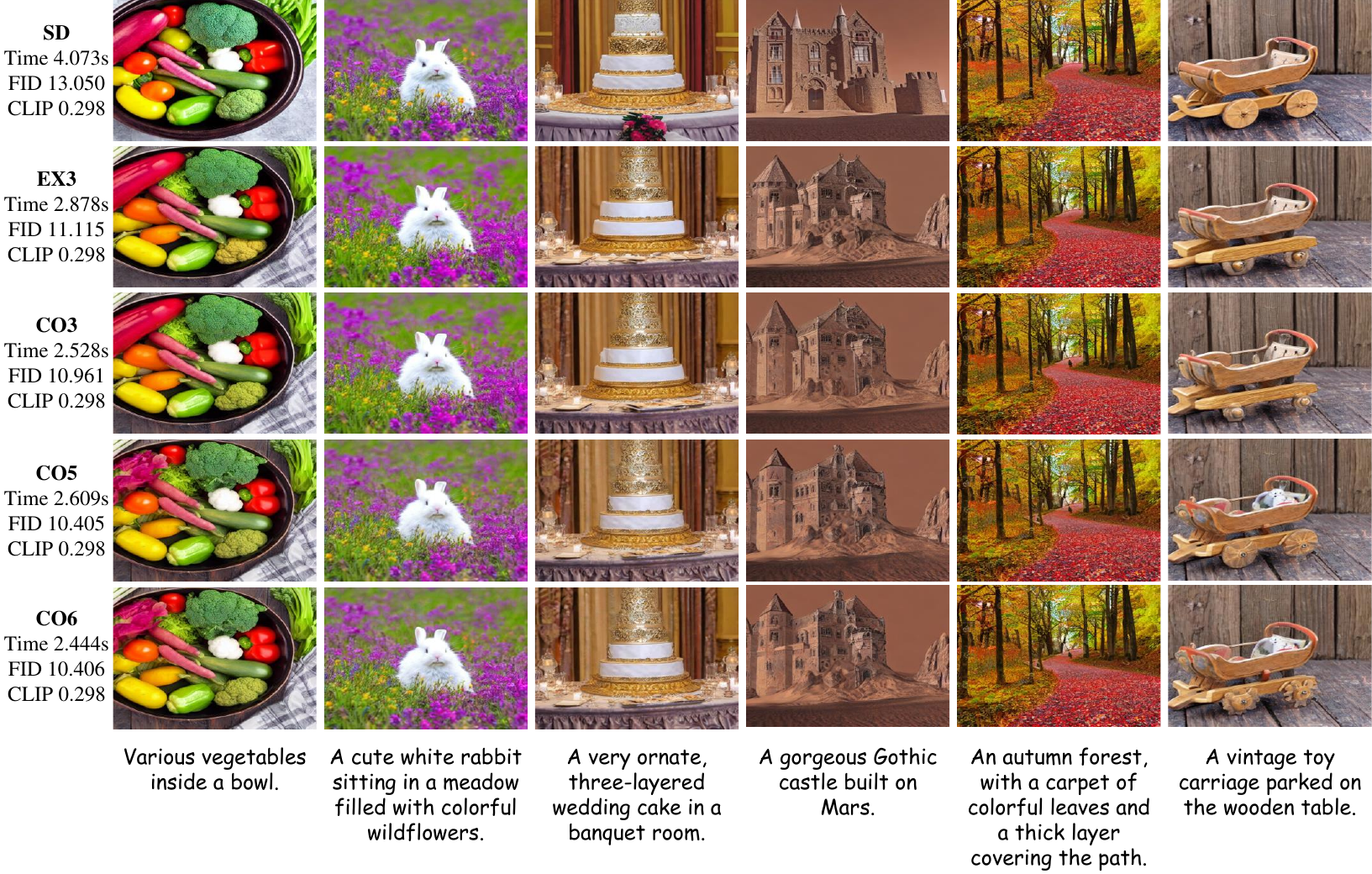}
	\caption{Comparison of feature inheritance strategies (EX3, CO3, CO5, CO6) in $P5\dag$ sampling mode to SDM without feature inheritance. The number of sampling steps is uniformly set to 25.
	}
	\label{14_SeeIheritance}
\end{figure*}
% unit-level
In addition to the layer-level computation skip mentioned earlier, we further introduce two \textit{unit-level} feature inheritance strategies to investigate the role of ResUnit and AttnUnit in the UNet.
The AttnUnit feature inheritance (AI) strategy skips the calculation of all attention units in the UNet.
However, this strategy significantly degrades network performance, indicating that attention units have a crucial impact on generation quality.
In contrast, the ResUnit feature inheritance (RI) strategies, which skip ResUnit calculations through feature inheritance, perform particularly well on the FID metric.
Based on these observations, we prioritize the study of AttnUnit in the concurrent feature inheritance part.

In the \textit{block-level} feature inheritance experiment, we observe that the external loop (EX1-3) and decoder loop (DE1-3)  strategies outperform the internal loop (IN1-3) and encoder loop (EN1-3) strategies significantly. This suggests that the decoder part and the shallow layers are critical in performance improvement.
Generally, the shallow layers of the UNet are primarily responsible for detail optimization, while the decoder part is responsible for expressing the generated content. Therefore, the experimental results align with this intuitive understanding.

Based on the experimental findings, we devised specific strategies for concurrent feature inheritance (CO1 and CO2) tailored to the decoder and shallow loop structures in the \textit{concurrent feature inheritance} part.
Additionally, we introduced the CO3 strategy for cyclic calculations within the up3 block. 
For the shallow layers, we further explored local unit loop strategies (CO4-6).
The experimental comparison highlights the significance of the dn0 and up3 blocks in enhancing FID indicators, particularly due to the role of AttnUnit within these blocks.
CO3, CO5, and CO6 demonstrate notable effectiveness.
Given the computational complexity of shallow attention, the CO6 strategy emerges as a promising feature inheritance structure, balancing speed and performance considerations.

In the $P5\dag$ sampling mode, Fig.~\ref{14_SeeIheritance} illustrates the generated results of feature inheritance strategies alongside non-inheritance approaches.
These results underscore the efficacy of feature inheritance strategies in improving computational efficiency while preserving the quality of generated outcomes.  % See \ref{appendix-c} for more samples.

\textbf{Sampling mode study:}
After comparing the differences in feature inheritance structures, we further delved into the impact of sampling modes. 
Tab.~\ref{FI-SOTA} compares the outcomes of $P5\dag$ and $P5$, revealing that the results of $P5\dag$, which excludes feature inheritance in the last 10 steps, generally outperform those of $P5$, which employs feature inheritance throughout the process.
This aligns with the conclusion drawn in Sec.~\ref{multi-unet}, emphasizing the importance of fully optimizing the final stage of the inference process with an original UNet to enhance generation quality.

Comparing CO3, CO5, CO6, LI1, and RI in $P5\dag$ and $P2\dag$, we observe that extending the feature inheritance period to 5 leads to a rapid reduction in the FID score and an acceleration in network computing speed.
This suggests that the shallow layer responsible for detail optimization has a significant impact on the FID index.
The superior-performing strategies in Tab.~\ref{FI-SOTA} typically skip the deep and encoder parts of the calculation, focusing on iteratively optimizing the shallow and decoder parts, particularly the attention units in these parts.
However, with a longer feature inheritance period, the calculation of the AttnUnit part across the entire network diminishes, leading to a slight decline in the CLIP index. Additionally, due to the reduction in the deep part calculation, the IS metric also exhibits a moderate decline.
We show the results of sampling mode $Pn$ ($n$ represents period) for different inheritance periods under the CO6 structure in \ref{appendix-c} , and it can be seen that the generation quality starts to decline as the period grows to 10.

In summary, compared with conventional 50-step and 25-step SDM generation methods, feature inheritance strategies such as CO6, CO5 and CO3 effectively improve inference speed and generation quality under $P5\dag$ sampling mode.

% \subsection{Comparison with State-of-the-art} 
% \textcolor{blue}{\textbf{Quantitative evaluation.}
	% xxx xxx xxx xxx xxx xxx xxx xxx xxx xxx xxx xxx xxx xxx xxx xxx xxx xxx xxx xxx xxx xxx xxx xxx xxx }  

\section{Conclusion}
In this paper, we tackle the issue of local computational redundancy in the diffusion models and then propose tuning and tuning-free methods to optimize the model based on our analysis.
For the tuning method, we introduce a model assembly strategy aimed at pruning redundant layers from the UNet of the SDM while preserving performance.
Additionally, to maintain performance in the minimal distillation model, we incorporate ME-CondConv in the pruning part to offset capacity loss resulting from pruning, thereby enhancing network performance and computational speed.
Furthermore, we explore the multi-UNet switching method to improve generation speed.
In the tuning-free method section, we propose a feature inheritance strategy enabling block-level, layer-level, and unit-level feature inheritance, to significantly accelerate generation.
We further investigate feature inheritance strategies under different sampling modes  from the perspective of time step.
% with different period strategies.
Experimental results demonstrate that the UNet speed of the lightweight model using the tuning model assembly strategy is $22.4\%$ faster than SDM.
%, with an overall speed 19\% faster than the original SDM. 
Moreover, the proposed feature inheritance strategy enhances the generation speed of SDM by $40.0\%$ without additional tuning.

\bibliographystyle{unsrt}% elsarticle-num before plain
\bibliography{ref}  % ref

% {\appendix[Proof of the Zonklar Equations]
	% Use $\backslash${\tt{appendix}} if you have a single appendix:
	% Do not use $\backslash${\tt{section}} anymore after $\backslash${\tt{appendix}}, only $\backslash${\tt{section*}}.
	% If you have multiple appendixes use $\backslash${\tt{appendices}} then use $\backslash${\tt{section}} to start each appendix.
	% You must declare a $\backslash${\tt{section}} before using any $\backslash${\tt{subsection}} or using $\backslash${\tt{label}} ($\backslash${\tt{appendices}} by itself
	%  starts a section numbered zero.)}

% {\appendices
	% \section*{Proof of the First Zonklar Equation}
	% Appendix one text goes here.
	% You can choose not to have a title for an appendix if you want by leaving the argument blank
	% \section*{Proof of the Second Zonklar Equation}
	% Appendix two text goes here.}

\clearpage

\appendices

\onecolumn
\begin{center}
	\LARGE \bfseries Supplementary Materials\\[1em]
\end{center}

\section*{Appendix A. Additional Results of A-SDM}
\label{appendix-a}
\begin{figure*}[h] %htbp
	\centering
	\includegraphics[width=1\textwidth]{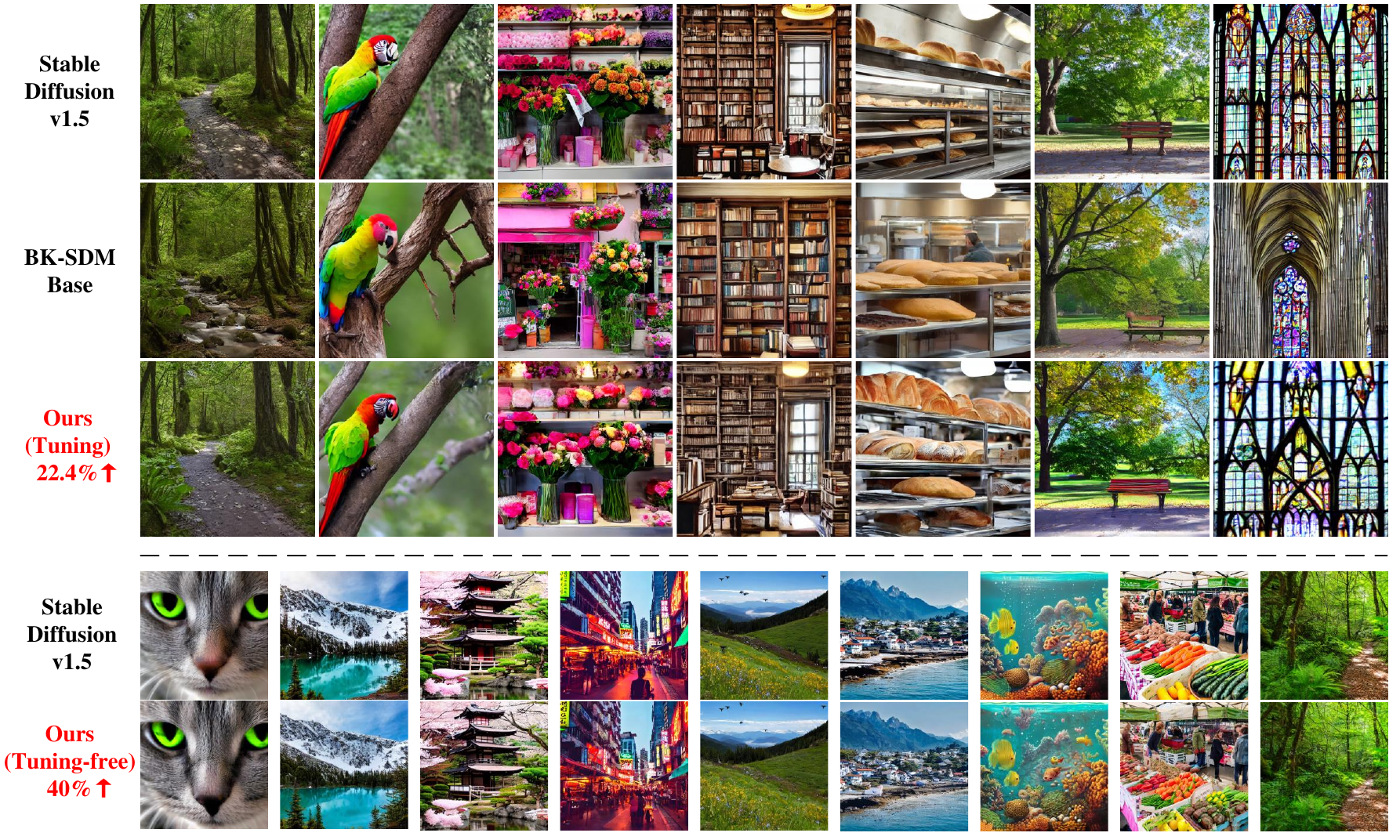}
	\caption{Accelerating Stable Diffusion v1.5 by 22.4\% and 40.0\% using tuning (model assembly) and tuning-free (feature inheritance) methods.} % , with 25 and 50 PLMS steps, respectively.
	\label{15_samples}
\end{figure*}
\newpage

\section*{Appendix B. Additional Results of Model Assembly Strategy}
\label{appendix-b}

\begin{figure*}[h]
	\centering
	\includegraphics[width=1\textwidth]{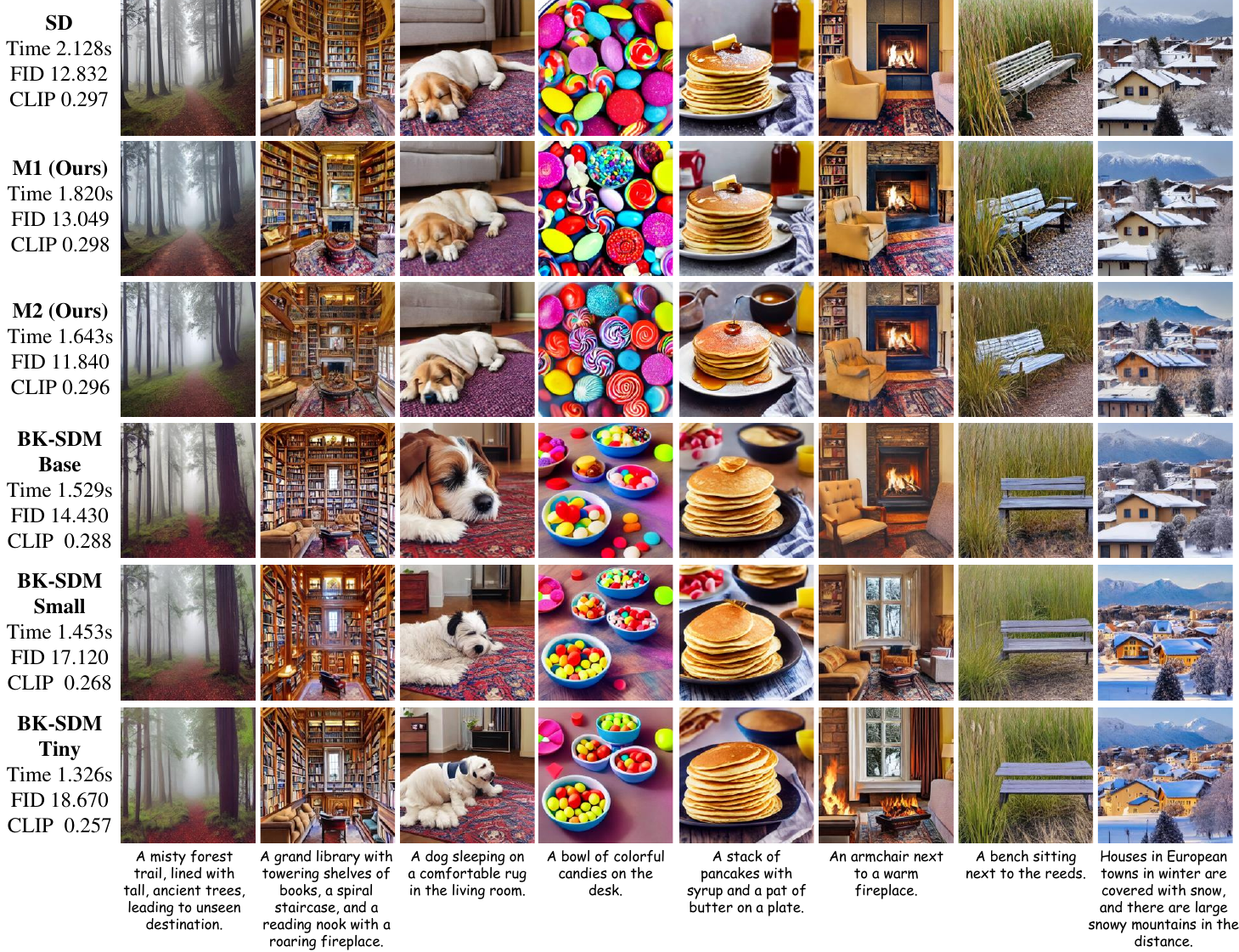}
	\caption{Visual comparison of outputs of SDM, M1, M2, Base, Small, and Tiny models with 25 sampling steps. The visual comparison shows that the proposed model assembly strategy (M1, M2) of deep partial freezing has good semantic stability and visual consistency with the results generated by the original SD model. While the Base, Small and tiny models show relatively large differences compared with the SDM model.}
	\label{16_BaseSmallTiny}
\end{figure*}

\newpage

\section*{Appendix C. Additional Results of Feature Inheritance Strategy}
\label{appendix-c}

\begin{figure*}[h]
	\centering
	\includegraphics[width=1\textwidth]{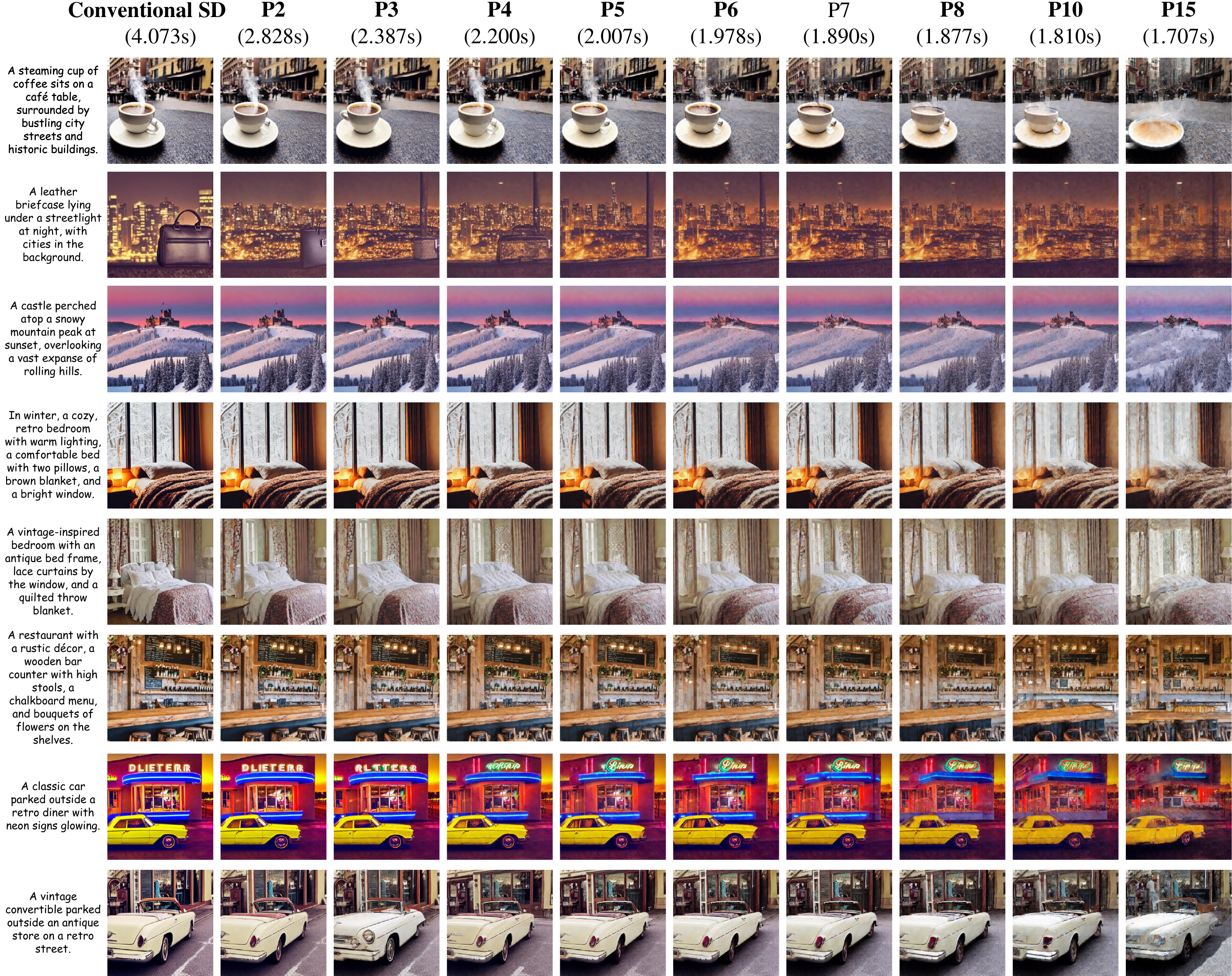}
	\caption{Additional results of concurrent feature inheritance strategy CO6 under different feature inheritance periods ($P2$-$P15$). In the $Pn$ sampling mode, when the period $n$ of feature inheritance is extended to 8-15, the image quality is significantly reduced.}
	\label{17_Pn}
\end{figure*}

\end{document}